%% file: main.tex
\documentclass[acmsmall]{acmart}

\AtBeginDocument{%
  \providecommand\BibTeX{{%
    \normalfont B\kern-0.5em{\scshape i\kern-0.25em b}\kern-0.8em\TeX}}}

\usepackage{multirow}
\usepackage{makecell}
\usepackage{xspace,mfirstuc,tabulary}
\usepackage{booktabs}
\usepackage{graphicx}
\usepackage{bbding}
\usepackage{xcolor,colortbl}
\usepackage{adjustbox}
\setcitestyle{numbers}
\usepackage{fontawesome5}
\usepackage{hyperref}
\usepackage{pifont}
\usepackage{xcolor}
\definecolor{hidden-black}{rgb}{0,0,0}
\newrobustcmd{\B}{\bfseries}
\newcommand*\colourcheck[1]{%
  \expandafter\newcommand\csname #1check\endcsname{\textcolor{#1}{\ding{52}}}%
}
\newcommand*\colourcross[1]{%
  \expandafter\newcommand\csname #1cross\endcsname{\textcolor{#1}{\ding{55}}}%
}
\colourcheck{green}
\colourcross{red}

\usepackage[most]{tcolorbox}
\newtcolorbox{highlighted}{
  colback=yellow!50!white, colframe=yellow!50!white, boxrule=0pt, sharp corners, left=0pt, right=0pt, top=0pt, bottom=0pt
}

\definecolor{cGreen}{RGB}{0,150,0} 
\newcommand{\cmark}{\textcolor{cGreen}{\ding{51}}} 
\newcommand{\xmark}{\textcolor{red}{\ding{55}}}    

\usepackage{color,soul}
\soulregister{\textcolor}{2}
\soulregister{\cite}7
\soulregister{\citep}7
\soulregister{\citet}7
\soulregister{\ref}7

\usepackage[switch]{lineno}
\sethlcolor{yellow}

\usepackage[edges]{forest}
\definecolor{lightcoral}{rgb}{0.94, 0.5, 0.5}
\definecolor{lightgreen}{rgb}{0.56, 0.93, 0.56}
\definecolor{harvestgold}{rgb}{0.98, 0.85, 0.40}
\definecolor{brightlavender}{rgb}{0.75, 0.58, 0.89}
\definecolor{capri}{rgb}{0.0, 0.75, 1.0}
\definecolor{carminepink}{rgb}{0.92, 0.3, 0.26}
\definecolor{celadon}{rgb}{0.67, 0.88, 0.69}
\definecolor{darkpastelgreen}{rgb}{0.01, 0.75, 0.24}

\definecolor{hidden-draw}{RGB}{205, 44, 36}
\definecolor{hidden-blue}{RGB}{194,232,247}
\definecolor{hidden-orange}{RGB}{243,202,120}
\definecolor{hidden-yellow}{RGB}{242,244,193}
\definecolor{tree-level-1}{RGB}{245,20,85}
\definecolor{tree-level-2}{RGB}{246,86,118}
\definecolor{tree-level-3}{RGB}{248,177,193}
\definecolor{tree-leaf}{RGB}{176,230,198}

\begin{document}

\title{AI Meets Brain: A Unified Survey on Memory Systems from Cognitive Neuroscience to Autonomous Agents}


\author{Jiafeng Liang}\authornote{Equal contribution}
\email{jfliang@ir.hit.edu.cn}
\author{Hao Li}\authornotemark[1]
\email{haoli@ir.hit.edu.cn}
\affiliation{%
	\institution{Harbin Institute of Technology}
	\city{Harbin}
	\state{Heilongjiang}
	\country{China}
	\postcode{150001}}

\author{Chang Li}\authornotemark[1]
\email{cli@ir.hit.edu.cn}
\author{Jiaqi Zhou}\authornotemark[1]
\email{yufeiqs917@gmail.com}
\affiliation{%
	\institution{Harbin Institute of Technology}
	\city{Harbin}
	\state{Heilongjiang}
	\country{China}
	\postcode{150001}}

\author{Shixin Jiang}
\email{sxjiang@ir.hit.edu.cn}
\author{Zekun Wang}
\email{zkwang@ir.hit.edu.cn}
\affiliation{%
	\institution{Harbin Institute of Technology}
	\city{Harbin}
	\state{Heilongjiang}
	\country{China}
	\postcode{150001}}

\author{Changkai Ji}
\email{ckji24@m.fudan.edu.cn}
\affiliation{%
 \institution{Fudan University}
 \city{Shanghai}
 \state{Shanghai}
 \country{China}
 \postcode{200433}}

\author{Zhihao Zhu}
\email{zhzhu@ir.hit.edu.cn}
\author{Runxuan Liu}
\email{rxliu@ir.hit.edu.cn}
\affiliation{%
	\institution{Harbin Institute of Technology}
	\city{Harbin}
	\state{Heilongjiang}
	\country{China}
	\postcode{150001}}

\author{Tao Ren}
\email{rtkenny@stu.pku.edu.cn}
\affiliation{%
	\institution{Peking University}
	\city{Beijing}
	\state{Beijing}
	\country{China}
	\postcode{100871}}

\author{Jinlan Fu}
\email{jinlanjonna@gmail.com}
\author{See-Kiong Ng}
\email{seekiong@nus.edu.sg}
\affiliation{%
	\institution{National University of Singapore}
	\city{Singapore}
	\state{Singapore}
	\country{Singapore}
	\postcode{119077}}

\author{Xia Liang}\authornote{Corresponding author}
\email{xia.liang@hit.edu.cn}
\author{Ming Liu}\authornotemark[2]
\email{mliu@ir.hit.edu.cn}
\author{Bing Qin}
\email{qinb@ir.hit.edu.cn}
\affiliation{%
	\institution{Harbin Institute of Technology}
	\city{Harbin}
	\state{Heilongjiang}
	\country{China}
	\postcode{150001}}

\renewcommand{\shortauthors}{Liang, et al.}

\begin{abstract}
Memory serves as the pivotal nexus bridging past and future, providing both humans and AI systems with invaluable concepts and experience to navigate complex tasks. 
Recent research on autonomous agents has increasingly focused on designing efficient memory workflows by drawing on cognitive neuroscience.
However, constrained by interdisciplinary barriers, existing works struggle to assimilate the essence of human memory mechanisms. 
To bridge this gap, this survey systematically synthesizes interdisciplinary knowledge of memory, connecting insights from cognitive neuroscience with LLM-driven agents. 
Specifically, we first elucidate the definition and function of memory along a progressive trajectory from cognitive neuroscience through LLMs to agents. 
We then provide a comparative analysis of memory taxonomy, storage mechanisms, and the complete management lifecycle from both biological and artificial perspectives. 
Subsequently, we review the mainstream benchmarks for evaluating agent memory. 
Additionally, we explore memory security from dual perspectives of attack and defense. 
Finally, we envision future research directions, with a focus on multimodal memory systems and skill acquisition.
\textcolor{blue}{Code: \href{https://github.com/AgentMemory/Huaman-Agent-Memory}{https://github.com/AgentMemory/Huaman-Agent-Memory}}
\end{abstract}





\maketitle

\input{1_Introduction}
\input{2_Definition}

\input{3_Utility}
\input{4_Taxonomy}
\input{5_Storage}
\input{6_Management}
\input{7_Benchmark}

\input{8_Security}

\input{9_Future}

\input{10_Conclusion}

\bibliographystyle{ACM-Reference-Format}
\bibliography{custom}


\end{document}

%% file: 1_Introduction.tex
\section{Introduction}

Memory is the cognitive nexus that interweaves past experience with future decision-making~\cite{brain_intro_memories}. 
In humans, it manifests as a dynamic neural process through which the brain stores and manages information~\cite{cog_brain_system}. 
Its profound significance lies in endowing individuals with the capacity to learn, adapt, and reshape their behavior, enabling humans to maintain coherence and foresight in an ever-changing environment~\cite{brain_intro_evolutionary,brain_intro_adaptive}.
As Large Language Models (LLMs) continue to evolve, endowing AI systems with human-like memory capabilities has emerged as a critical challenge~\cite{summary_ltm}. 
However, the natively stateless nature~\cite{survey_stateless} of LLMs renders each inference independent, preventing models from maintaining cross-session continuity or accumulating experience from historical interactions.
While modern LLMs have scaled parameters and context windows to massive sizes, knowledge update costs~\cite{summary_cost} and computational complexity~\cite{method_lost_in_middle,intro_on2} remain significant bottlenecks. 

With the rapid advancement of agents across diverse domains and tasks~\cite{method_cellagent,method_branch_and_browse,method_coge,method_memorb,method_sweexp,method_liddia,method_magis,method_experepair,method_fkn,method_recmind,method_toolmem}, memory systems have emerged as a critical factor in enhancing their performance by enabling information persistence~\cite{method_rgmem,method_pre_storage,method_pisa} and long-horizon planning~\cite{method_legomem,method_reasoningbank}.
Rather than serving as a passive repository for historical interactions, memory has evolved into a dynamic cognitive hub that underpins complex decision-making. 
However, despite significant progress in memory mechanism research in recent years, existing works~\cite{method_mem0,dy_memory_amem,method_mem1,method_memgpt,method_bot} tend to remain confined to a single disciplinary perspective or lack depth in biological research, making it difficult to achieve deep integration between cognitive science and artificial intelligence. 
This separation has deprived both fields of opportunities for deep mutual validation and inspiration in memory research.

To bridge this gap, our survey provides a comprehensive and unified review of memory systems, integrating insights from cognitive neuroscience with the rapidly evolving field of LLM-driven agents. 
We first establish a progressive research perspective on memory, transitioning from human brain to LLMs and ultimately to agents, systematically elucidating its definition and fundamental role (\S\ref{sec:Definitions}, \S\ref{sec:Utility}).
Building on the classical short- and long-term memory dichotomy in cognitive neuroscience, we propose a taxonomy that classifies agent memory along two dimensions, including nature- and scope-based classification (\S\ref{sec:Categorization}).
The former distinguishes procedural experience from conceptual knowledge, while the latter concerns memory persistence within or across trajectories.
Next, we examine memory storage from the perspectives of location and format (\S\ref{sec:Storage}).
In cognitive neuroscience, short-term memory relies on distributed sensory-frontoparietal networks while long-term memory depends on hippocampal-neocortical coordination. 
For agents, storage locations include the context window for temporary memory and external memory bank for persistent information. 
Regarding format, the brain employs persistent activity and synaptic connection weights for short-term retention and structured forms like cognitive maps for long-term memory, while agents utilize natural language text, graph structures preserving relational information, internalized parameters, and latent representation in high-dimensional vector spaces.

This survey further analyzes memory management mechanisms in both the human brain and agents, covering the complete closed-loop lifecycle of memory extraction, updating, retrieval, and utilization (\S\ref{sec:Management}). 
In cognitive neuroscience, new information is encoded and gradually stabilized into durable representations through hippocampal-neocortical coordination. 
When external cues trigger hippocampal replay, the information about past events carried by these representations is reinstated, and the retrieval process itself opens a plasticity window during which the underlying memory traces can be updated, strengthened, or weakened.
In agents, raw information is distilled into structured records through flat, hierarchical, or generative paradigms, dynamically refreshed within trajectories while maintained across them. 
These records are retrieved via similarity matching or multi-factor approaches, then incorporated into reasoning through contextual augmentation or parameter internalization.
Then, we comprehensively outline various benchmarks for evaluating agent memory capabilities, categorizing them into semantic-oriented benchmarks that examine internal state maintenance and higher-order cognitive abilities, and episodic-oriented benchmarks that assess performance in vertical domains such as web search, tool use, and environmental interaction (\S\ref{sec:Benchmarks for Agent Memory}).
Furthermore, we address the often-overlooked yet critical issue of memory security, providing a systematic investigation from both attack and defense perspectives (\S\ref{sec:Security}). 
On the attack side, existing methods focus on extracting sensitive information, implanting backdoors through malicious data, or degrading agent judgment by introducing noise and conflicting signals. 
On the defense side, countermeasures have been developed to purify retrieval sources, block harmful responses in real time, and safeguard sensitive data throughout the memory lifecycle.
Finally, we propose future research directions with particular focus on two key areas (\S\ref{sec:Future}). 
The first is multimodal memory systems capable of processing and integrating information across text, image, audio, video modalities. 
The second is agent skills that enable memory sharing and transfer across heterogeneous agents, transforming domain expertise into composable, reusable, and portable modular resources.
We hope this survey can facilitate cross-disciplinary research, offering newcomers an accessible introduction while providing seasoned researchers with insights.

%% file: 2_Definition.tex
\section{Memory Definitions}
\label{sec:Definitions}

To establish a comprehensive understanding of memory, our exposition systematically progresses from cognitive neuroscience to autonomous agents. 
We commence by examining the foundational principles of memory from a cognitive neuroscience perspective (\S\ref{ssec:Memory from the Perspective of Cognitive Neuroscience}). 
Subsequently, we transition to the perspective of Large Language Models (LLMs), elucidating the manifestations of memory within this paradigm (\S\ref{ssec:Memory from the Perspective of Large Language Models}). 
Finally, we explore the perspective of agents, investigating how memory operates as a dynamic cognitive mechanism within agent architectures to facilitate sophisticated task planning and environmental interaction (\S\ref{ssec:Memory from the Perspective of Agents}).

\subsection{Memory from the Perspective of Cognitive Neuroscience}
\label{ssec:Memory from the Perspective of Cognitive Neuroscience}
Memory is fundamentally a process through which the brain processes and manages information.
This process can be roughly divided into two stages. 
In the first stage, as the brain acquires new concepts or encounters novel events, it rapidly forms specific neural representations while simultaneously encoding and integrating this information.
In the second stage, the brain operates on stored representations, either consolidating them over time or retrieving them in response to similar future situations.
Moreover, memory can exert lasting effects on cognition and behavior, influencing how people think and act both in the present~\cite{brain_what_is_memory_2} and in the future~\cite{brain_what_is_memory_1}. 
In this sense, memory serves as a cognitive bridge connecting past experiences with future decisions, rather than merely storing and mechanically replaying information~\cite{what_memory_is}. 
It is precisely this capacity that makes memory the most fundamental and core capability of intelligence. 

\subsection{Memory from the Perspective of Large Language Models}
\label{ssec:Memory from the Perspective of Large Language Models}
In LLMs, memory does not exist as a monolithic storage construct, but rather manifests in multiple distinct forms spanning different storage substrates. 
These diverse memory modalities are essential for overcoming the natively stateless~\cite{survey_stateless} generative nature of such models and enabling complex logical reasoning and multi-turn dialogue.
Contemporary research categorizes LLMs memory into three core types: 
(1) Parametric memory (\S\ref{sssec:Parametric Memory}), which is internalized within model parameters,
(2) Working memory (\S\ref{sssec:Working Memory}), which relies on the context window for real-time interaction,
and (3) Explicit external memory (\S\ref{sssec:Explicit External Memory}), which is realized through external storage and retrieval mechanisms. 

\subsubsection{Parametric Memory}
\label{sssec:Parametric Memory}
Parameter memory refers to parameterized knowledge encoded within the weights of a neural network~\cite{para_memory_train}. 
This constitutes the most fundamental form of memory in LLMs and serves as the cornerstone of their intelligent capabilities. 
From a cognitive neuroscience perspective, this modality corresponds to abstract long-term memory representations in human cognition.
Such memories are acquired during the pre-training and post-training phases, wherein patterns extracted from vast corpus are compressed and internalized into high-dimensional parametric representations. 
This process endows the model with robust generalization capabilities and establishes the foundation for logical reasoning.
However, its static and encapsulated storage characteristics impose significant limitations.
First, there exists an inherent temporal lag, as parameters become frozen upon completion of training, rendering the model incapable of perceiving information beyond the knowledge cutoff point~\cite{para_memory_para}.
Second, there is the phenomenon of factual hallucination~\cite{survey_huanglei}, as knowledge is encoded as probabilistic distributions rather than precise factual records, rendering the model susceptible to generating erroneous information in the absence of definitive evidence~\cite{para_memory_hall}.
Furthermore, its high training computational costs and potential catastrophic forgetting risk make it difficult to support the demands of highly reliable dynamic applications~\cite{para_memory_zero}.

\subsubsection{Working Memory}
\label{sssec:Working Memory}
In the cognitive architecture of LLMs, working memory is primarily based on the context window as its core carrier, performing the function of interaction and reasoning in real-time~\cite{working_memory_window}. 
Mechanistically, it refers to the length of the input sequence that the model can simultaneously perceive and process in a single reasoning iteration, directly determining the information throughput of the model's context learning. 
At the underlying implementation level, working memory mainly manifests as key-value caches of attention~\cite{working_memory_trans}. 
As the number of dialog turns increases, the model needs to maintain and update this cache state in real time to maintain contextual coherence. 
However, this mechanism faces significant physical limitations. 
Firstly, there is a contradiction between capacity and cost. 
Although state-of-the-art models like Gemini 3~\cite{working_memory_gemini} and GPT-5~\cite{working_memory_chat} have successfully extended context windows to the scale of millions of tokens, the high computational cost of processing such long sequences still limits their practicality as the primary carriers for storing massive amounts of information.
Furthermore, the context window exhibits a positional bias similar to human memory. 
The model tends to pay more attention to the first and last pieces of information in the window~\cite{working_memory_gate,working_memory_sink} and is more likely to ignore the middle sections~\cite{working_memory_middle}.
Therefore, the context window is more suitable for handling high-frequency, immediate task instructions than serving as a long-term, stable knowledge base.

\subsubsection{Explicit External Memory}
\label{sssec:Explicit External Memory}
To transcend the static boundaries of parametric storage and physical constraints of context windows, explicit extended memory introduces a storage medium independent of neural network weights, thereby establishing a non-parametric knowledge augmentation mechanism~\cite{method_hipporag}. 
The core design philosophy underlying this approach is the decoupling of computation and storage, wherein LLMs function as a central processing unit responsible for inference and scheduling, while vast quantities of knowledge are offloaded to external repositories. 
This architectural shift fundamentally transforms the model from a passive knowledge repository into an active knowledge orchestrator~\cite{ex_memory_km}. 
The prevailing paradigm in this domain is retrieval-augmented generation (RAG)~\cite{ex_memory_rag}, which leverages vector databases~\cite{ex_memory_vd} or knowledge graphs~\cite{ex_memory_kg} to store extensive volumes of information.
Such system effectively mitigate the hallucination phenomena and temporal lag inherent in parametric memory, enabling real-time knowledge updates and precise provenance tracking at minimal computational cost, thereby substantially enhancing the reliability of system outputs.
However, inference latency induced by retrieval operations, noise interference from irrelevant contexts, and the complexity of large-scale index construction under this paradigm remain critical bottlenecks constraining its performance~\cite{ex_memory_survey}.

\subsection{Memory from the Perspective of Agents}
\label{ssec:Memory from the Perspective of Agents}
From the perspective of autonomous agents, the conceptualization of memory transcends the mere data storage paradigm characteristic of LLM-centric approaches, evolving into a sophisticated dynamic cognitive architecture~\cite{agent_memory_adv}.
Specifically, the LLM-centric perspective primarily alleviates the physical constraints imposed by parameters and context windows, while the agentic perspective shifts focus toward leveraging neuroscientific principles to construct external memory systems.
These systems augment the foundational reasoning capabilities of agents by endowing them with capacities for identity persistence, experiential accumulation, and long-horizon planning~\cite{agent_memory_see}.
To elucidate these complex operational principles, this section diverges from the storage medium discourse in LLMs perspective, opting instead to deconstruct memory along three core dimensions: 
(1) Structured storage (\S\ref{sssec:Structured Storage}), 
(2) Dynamic scheduling mechanisms (\S\ref{sssec:Dynamic Scheduling Mechanisms}), 
and Cognitive processing and evolution (\S\ref{sssec:Cognitive Processing and Evolution}).
Additionally, to provide a clearer introduction and prevent confusion, we compare the differences between agent memory and RAG (\S\ref{sssec:The Difference Between Agent Memory and RAG}).

\subsubsection{Structured Storage}
\label{sssec:Structured Storage}
Structured storage serves as the physical carrier of agent memory systems, aiming to transform unstructured natural language interactions into an efficient format that is easy for machines to index and understand~\cite{str_memory_os}. 
Unlike the implicit parameter distribution within LLMs, agent external memory typically employs heterogeneous hybrid storage strategies. 
It mainly exhibits three typical paradigms, namely temporal flow, hierarchical flow, and symbolic libraries. 
Specifically, \citet{str_memory_gen} constructed a linear memory that encapsulates interaction records as discrete memory objects, including timestamps, text descriptions, and importance scores.
To overcome the physical boundaries of long-range contexts, \citet{str_memory_walk} adopted a hierarchical memory tree structure, constructing a pyramid-like index through upward recursive summarization, enabling agents to locate key information through navigation rather than complete processing. 
In procedural memory, \citet{str_memory_voya} introduced a skill library similar to key-value pairs, using semantic vectors of skill descriptions as keys and executable code as values, successfully transforming unstructured exploration experiences into structured, reusable, and composable executable programs. 
These structural evolutions demonstrate that the memory system has transformed from a static data container into a dynamic cognitive center supporting complex decision-making.

\subsubsection{Dynamic Scheduling Mechanisms}
\label{sssec:Dynamic Scheduling Mechanisms}
While structured storage addresses the challenge of information persistence, dynamic scheduling mechanisms tackle the tension between limited attention resources and vast memory stores~\cite{dy_memory_mul}.
Given that LLMs are constrained by the physical limitation of context windows, agents cannot input all historical information into the model at once.
Therefore, it is necessary to establish an efficient memory retrieval and update mechanism to achieve accurate routing and dynamic adaptation of information flow in the memory system. 
Regarding the screening of redundancy mechanisms, \citet{dy_memory_light} effectively balanced memory coverage and maintenance latency through lightweight compression of sensory memory and updated strategy during dormancy. 
This hierarchical idea is also very important in multi-agent collaboration.
\citet{dy_memory_co} proposed a time-sharing scheduling strategy that borrows from the process scheduling of the operating system and designed a memory model based on time slicing and multi-level caching, which significantly improved the resource allocation efficiency in a concurrent environment. 
In addition to static hierarchical and scheduling mechanisms, learning-based adaptive mechanisms are introduced to cope with more dynamic and complex environmental changes. 
The improved IGP-PPO algorithm~\cite{dy_memory_dy} was employed to enable adaptive decision-making by dynamically selecting the optimal scheduling rule based on real-time states.
\citet{dy_memory_mem} applied reinforcement learning to long-range text stream management, training agents to autonomously retain and overwrite memory, thereby achieving information extrapolation with linear complexity.
Based on this, the memory scheduling mechanism further evolved towards autonomous evolution.
\citet{dy_memory_amem} proposed a dynamic mechanism that can autonomously trigger Link Generation and memory evolution, and by updating the context description of existing memory and reconstructing semantic associations in real time during interaction, it achieved the self-growth of knowledge networks without predefined rules. 
The dynamic scheduling mechanism successfully broke through physical bottlenecks through more refined resource management and adaptive strategies, and ensured the efficient flow of interactive information.

\subsubsection{Cognitive Processing and Evolution}
\label{sssec:Cognitive Processing and Evolution}
Dynamic scheduling of memory solves the problem of precise location, but it is difficult to summarize and generalize from fragmented raw record to general experience~\cite{cog_memory_agentworkflow}. 
To achieve this, agents have to deeply reflecting on, abstracting, and reorganizing memory content, thereby driving continuous updates to their behavioral strategies~\cite{cog_memory_survey}.
First, the most direct form of cognitive processing is feedback-based self-reflection, where the agent optimizes future decisions by evaluating its historical behavior. 
Reflexion~\cite{cog_memory_relfex} abandoned the traditional model weight update and instead guided the agents to reflect on itself and store it in contextual memory through a language feedback mechanism, achieving the flexibility of learning from mistakes. 
On this basis, ExpeL~\cite{cog_memory_ex} enhanced the abstraction of learning by extracting insights from training tasks and directly calling these processed insights rather than raw data during reasoning. ReasoningBank~\cite{method_reasoningbank} further deepened the dimension of reflection by combining memory-aware test-time scaling to synthesize high-quality reasoning strategies from successful and unsuccessful experiences, thereby achieving the self-evolution of reasoning ability. 
Secondly, to overcome the limitations of specific tasks, the agents also need to have the ability to generalize general knowledge from specific experiences. SEDM~\cite{cog_memory_sedm} introduced cross-domain knowledge diffusion, which used lightweight abstraction operators to convert knowledge of a specific domain into a general form that removed the domain-specific details. 
This mechanism ensures that the source knowledge domain can be safely transferred and reused in new scenario tasks, thereby achieving efficient cross-domain memory sharing. 
Finally, to address the memory pool expansion and noise accumulation caused by long-term interactions, agents must possess forgetting and dynamic evolution mechanisms similar to those in humans to maintain the efficiency and relevance of memory.
SAGE~\cite{cog_memory_sage} borrowed from the Ebbinghaus forgetting curve~\cite{ebbinghaus} and significantly optimized processing capabilities in contextual scenarios by adaptively eliminating low-value information to reduce cognitive load. 
Regarding storage, MEMORYLLM~\cite{cog_memory_memllm} explored a scheme to construct a fixed-size memory pool in the latent space, balancing new and old knowledge while suppressing disordered memory growth, thus ensuring the continued effectiveness of the memory.

\subsubsection{The Difference Between Agent Memory and RAG}
\label{sssec:The Difference Between Agent Memory and RAG}
In terms of management mechanisms, there exists a significant similarity between agent memory mechanisms and retrieval-augmented generation (RAG) techniques, as both enable language models to transcend the limitations of their inherent parametric knowledge through extracting and utilizing external information resources~\cite{survey_rag,survey_agentmemory}. 
Although they converge in technical implementation, they exhibit distinct differences in usage scenarios. 
Traditional RAG approaches focus on connecting LLMs to static knowledge resources (e.g., Wikipedia, document library) for instant querying~\cite{ex_memory_rag,method_mainrag,method_mmed_rag}. 
The design objective of such systems lies in ensuring that generated content is grounded in reliable factual information, reducing the probability of models producing erroneous information or hallucinations~\cite{survey_huanglei,survey_hal_1}. 
However, they generally lack the capability to record and accumulate historical interaction information~\cite{method_memgpt}. 
In contrast, agent memory systems are embedded within the dynamic interaction process between agents and their environments, continuously incorporating information generated from agent operations and environmental feedback into memory containers~\cite{method_reasoningbank}.

%% file: 3_Utility.tex
\section{Memory Utility}
\label{sec:Utility}

In cognitive neuroscience, memory constitutes the neural processes through which the brain encodes, stores, and retrieves information, enabling individuals to retain past experiences and leverage them to guide ongoing behavior and inform future decision-making~\cite{brain_intro_memories,brain_what_is_memory_2}.
In LLM-driven agents, a fundamental tension exists between the inherent statelessness of models coupled with engineering constraints and the continuity required for complex, long-horizon tasks. 
Consequently, memory transcends its role as a mere passive repository bridging historical interactions and instead serves as a pivotal active component within the cognitive architecture of agents.
Concretely, the incorporation of memory fundamentally extends agent capabilities through three paradigms: 
(1) It mitigates the context window burden by utilizing structured information management to bypass the computational costs and attentional degradation inherent in long-context reasoning (\S\ref{ssec:Breaking Context Window Constraints}). 
(2) It facilitates deep personalization by supporting both immediate conversational coherence and the construction of enduring user profiles (\S\ref{ssec:Constructing Long-term Personalized Profiles}). 
(3) It empowers reasoning enhancement and autonomous evolution by creating a feedback loop that synthesizes historical experience with reflection and planning modules (\S\ref{ssec:Driving Experience-based Reasoning}). 
The following sections will rigorously examine these three functional utilities.

\begin{figure*}[th]
    \centering
    \includegraphics[width=\linewidth]{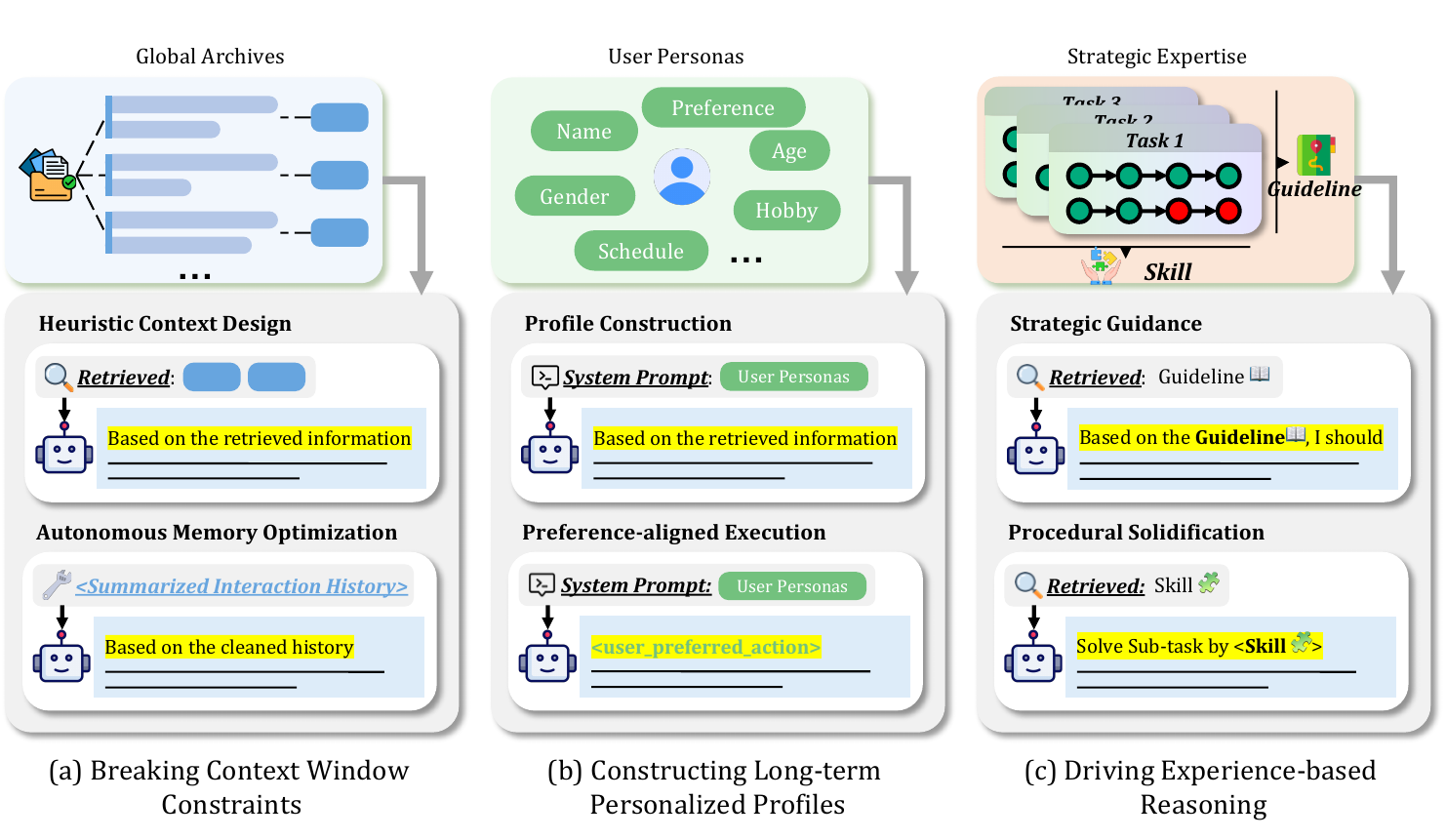}
    \caption{Overview of memory utility in LLM-driven agents. Memory extends agent capabilities by alleviating context window constraints, enabling long-term personalization, and driving experience-based reasoning through a feedback loop with reflection and planning.}
    \label{fig:utility}
\end{figure*}

\subsection{Breaking Context Window Constraints}
\label{ssec:Breaking Context Window Constraints}
Despite the expanding context windows of modern LLMs, the quadratic computational complexity of attention mechanisms and the lost-in-the-middle phenomenon~\cite{method_lost_in_middle} remain significant physical bottlenecks in long-horizon interactions. 
Consequently, the primary utility of memory lies in mapping infinite interaction streams into limited attention budgets, with the core paradigm shifting from passive linear truncation to dynamic context reconstruction. 
Depending on the implementation mechanism, this process manifests primarily through the utilization of hierarchical structural designs for physical compression and virtualization indexing, as well as the internalization of memory management as intrinsic agent actions to achieve end-to-end autonomous optimization.
Depending on the implementation mechanism, this process diverges into two primary directions: 
(1) Heuristic context design (\S\ref{ssec:Heuristic Context Design}), which utilizes hierarchical structural designs for physical compression and virtualization indexing, 
and (2) Autonomous memory optimization (\S\ref{ssec:Autonomous Memory Optimization}), which internalizes memory management as intrinsic agent actions to achieve end-to-end autonomous optimization, as shown in \autoref{fig:utility} (a).

\subsubsection{Heuristic Context Design}
\label{ssec:Heuristic Context Design}
Approaches rooted in heuristic context design aim to preserve global threads while concealing local details by drawing upon organizational methods from operating systems or human cognition~\cite{method_psychogat, method_graphcogent, method_cosmir, method_comorag, method_webresearcher}. 
\citet{method_memgpt} drew on operating system design principles to introduce virtual context management, utilizing paging mechanisms to bifurcate memory into immediate main context and archival external context. 
By scheduling information between these tiers via system instructions, it maintains the illusion of infinite context within constrained windows. 
To further enhance information density, \citet{method_readagent} mimicked human cognitive reading processes by proposing a gist memory mechanism, which paginates text and generates compressed gists as a global index, retrieving local raw text only when necessary. 
In dynamic interaction scenarios, \citet{method_foldgrpo} introduced a task-structure-based folding strategy, defining a main thread and branches to delimit information scopes and prevent sub-task details from polluting the main context. 
Complementarily, \citet{method_resum} addressed long-horizon search tasks with an iterative summarization mechanism, ensuring the preservation of critical navigation states during context compression.

\subsubsection{Autonomous Memory Optimization}
\label{ssec:Autonomous Memory Optimization}
Transcending the limitations of fixed heuristics, the latest research trend focuses on autonomous memory optimization by elevating memory management to a learnable intrinsic capability.  
\citet{method_supo,method_memsearcher} established the theoretical viability of this direction by demonstrating that summarization policies can be end-to-end optimized via reinforcement learning to maximize long-term rewards. 
Building on this, \citet{method_memory_as_action} pioneered the integration of summarization and deletion into the action space, allowing models to learn active memory management akin to tool usage. 
Further refining control granularity, \citet{method_agentfold} transformed context management into an explicit skill via a context folding mechanism. 
By issuing directives to toggle between granular condensation for high-resolution details and deep consolidation for abstract merging, it enabled agents to actively restructure their history, explicitly deciding which spans to preserve or compress based on task relevance.
This mechanism empowers agents to autonomously select compression granularity based on information salience, significantly reducing the loss of critical details in long-horizon web interactions and marking a complete transformation of memory mechanisms from passive storage to autonomous cognition.

\subsection{Constructing Long-term Personalized Profiles}
\label{ssec:Constructing Long-term Personalized Profiles}
The pre-training paradigm of LLMs fundamentally characterizes them as generalized knowledge engines, relying on fixed parametric knowledge to handle broad, common-sense tasks. 
However, this mechanism results in significant statelessness and homogeneity, leading to a one-size-fits-all interaction that fails to satisfy specific user preferences or meet the profound demand for personalized experiences in long-term interactions. 
To bridge this gap, memory mechanisms serve as a critical non-parametric complement within the agent architecture. 
Beyond the mere storage of historical interactions, their core utility lies in constructing a dynamically evolving cognitive model for the agent.
Through memory, agents can transform fragmented interaction data into structured user profiles, thereby achieving deep adaptation to users across two dimensions: 
(1) Profile construction (\S\ref{sssec:Profile Construction}) 
and (2) Preference-aligned execution (\S\ref{sssec:Preference-aligned Execution}),
as shown in \autoref{fig:utility} (b).

\subsubsection{Profile Construction}
\label{sssec:Profile Construction}
In the dimension of cognitive understanding, memory mechanisms endow agents with the capability to distill core traits from complex interaction streams and construct increasingly refined user models~\cite{method_macrs, method_memocrs,method_carmem,method_data_efficient, method_caim, method_bridging_long_term_gap, method_persona_aware_framework, method_interpersonal_memory, method_imperfect, method_prime, method_enabling_personalized,method_memqa,summary_the_personalization}. 
Works such as~\cite{method_generative_agents} have demonstrated that agents can act not merely as recorders but as reflectors, periodically reviewing memory streams to generate high-level insights and infer users' underlying motivations and personality traits.
To enhance the precision of profiling, frameworks like~\cite{method_secom} and~\cite{method_rmm} optimized memory utilization, ensuring agents can precisely recall the most relevant user background and preferences based on the current conversational context.
Building on this, \citet{method_memgpt,method_ld_agent} further reinforced structured profile maintenance by explicitly storing user attributes (e.g., name, profession) and bidirectional relationship states. 
This ensures that the agent maintains a coherent cognition of “who the user is” and “how the relationship stands” throughout long-horizon interactions, thereby establishing a robust and anthropomorphic emotional bond.

\subsubsection{Preference-aligned Execution}
\label{sssec:Preference-aligned Execution}
Building upon precise user profiles, the utility of memory mechanisms extends to the dimension of personalized decision-making, where historical experience guides concrete behavioral planning. 
As agents evolve into web agents capable of executing tasks, memory serves as an experiential guide for acting according to user habits. 
\citet{method_puma} demonstrated how memory banks act as a reference frame for implicit preferences, guiding agents to make decisions aligned with specific user habits during complex web task planning.
Complementarily, \citet{method_personaagent} showed that such adaptation can occur purely at test time, utilizing retrieved interaction patterns to construct dynamic personas that guide generation without parameter updates. 
In this paradigm, memory functions not merely as background knowledge but as a supervisory signal or dynamic constraint in the preference alignment process, enabling agents to predict and execute operations most aligned with user expectations, thereby elevating personalized services from mere verbal communication to the level of concrete task execution.

\subsection{Driving Experience-based Reasoning}
\label{ssec:Driving Experience-based Reasoning}
In the default setting of LLMs, each task is an isolated attempt, often leading to inefficient cycles of repetitive trial-and-error in long-horizon planning. 
The introduction of memory mechanisms bridges this cognitive gap, transforming the agent from a static solver into a continuous learner capable of drawing wisdom from history. 
By retaining successful trajectories and lessons from failures, memory endows agents with experience-based reasoning. 
Depending on the mode of intervention, this capability is realized through two primary paradigms: 
(1) Strategic guidance (\S\ref{sssec:Strategic Guidance}) assists the model in making superior plan by retrieving relevant experiences, guidelines, or optimized strategies
(2) Procedural solidification (\S\ref{sssec:Procedural Solidification}) solidifies high-frequency successful paths into workflows, code, or templates, enabling agents to bypass cumbersome planning processes and efficiently complete tasks by directly invoking procedural knowledge, as shown in \autoref{fig:utility} (c).

\subsubsection{Strategic Guidance}
\label{sssec:Strategic Guidance}
The first paradigm focuses on utilizing memory to assist reasoning, guiding the model to generate correct actions by providing high-quality contextual references or internalized strategic intuition~\cite{method_proagent, method_aga, cog_memory_ex, method_shieldagent, method_moba, method_dynamic_cheatsheet, method_rap, method_principle, method_mapagent, method_coarse_to_fine}. 
Early pioneering work \cite{method_reflexion} introduced textual reflection mechanisms, where agents convert failure experiences into linguistic constraints that serve a corrective function in subsequent reasoning. 
Following this, \citet{method_synapse} utilized memory to retrieve similar historical trajectories as exemplars, directly prompting the model to mimic past successful steps. 
As research advanced, this reference mechanism became more structured and dynamic. 
For instance, \citet{method_crmweaver} assisted problem-solving in business scenarios by retrieving high-similarity guidelines
If existing experience is insufficient, it employed a stronger model to generate new guidelines that are actively populated back into memory. 
Recent studies further employ memory for deep policy optimization.
\citet{method_tfgrpo} summarized sampled trajectory groups to extract advantageous experiences explaining why one trajectory succeeds while another fails, thereby guiding the optimal action selection during inference.
\cite{method_early_experience} transformed experience into internalized intuition by contrasting expert and non-expert actions from early exploration and combining them with implicit world model predictions.

\subsubsection{Procedural Solidification}
\label{sssec:Procedural Solidification}
The second paradigm focuses on procedural solidification, transforming successful reasoning processes into executable structures or skills to replace scratch-pad planning in subsequent tasks~\cite{method_os_copilot, method_skillweaver, method_archpilot}. 
\citet{str_memory_voya} pioneered this potential by encapsulating successful code generated during exploration into a skill library, allowing agents to invoke functions directly rather than generating action sequences when facing similar sub-tasks. 
\citet{method_cer} further developed the concept of skill abstraction through trajectory distillation, refining raw interactions into generalizable skills and environmental dynamics, which are replayed in context to enhance capabilities in new tasks. 
For more complex task processes, \citet{cog_memory_agentworkflow} captured critical patterns from historical trajectories while filtering out extraneous details, distilling generalizable workflow graphs through either offline extraction or online induction to serve as structured scaffolds for guiding subsequent actions.
To balance efficiency and safety, the recently proposed method~\cite{method_agentrr} designed a hierarchical experience abstraction mechanism that converts trajectories into multi-layer experiences, comprising high-level process knowledge and low-level executable action templates, while generating check functions for each layer. 
During the replay phase, the agent adaptively selects the appropriate experience hierarchy and instantiates specific actions, which substantially reduces reasoning overhead while ensuring operational determinism and transferability through structured templates.

%% file: 4_Taxonomy.tex
\section{Memory Categorization}
\label{sec:Categorization}

The concept of memory originally stems from cognitive neuroscience, where it is broadly defined as the cognitive process by which the brain stores and manage information, including experiences, facts, and skills, allowing this information to be accessed and used after the original stimulus or event is no longer present, and is typically classified into short- and long-term memory (\S\ref{sssec:Memory Classification in Cognitive Neuroscience}).

In LLM-driven agents, memory includes system's persistent past interaction information, which encompasses both perceptual records of environmental states and the history of interactions with the environment~\cite{method_react,method_reflexion}. 
Through the memory, agents can not only maintain contextual coherence and information consistency across multi-turn dialogues, but also extract valuable patterns and knowledge from historical experiences, thereby demonstrating adaptive learning and continuous improvement capabilities in task execution.
Traditionally, memory in agent domains is divided into short- and long-term memory~\cite{survey_beihang,method_agentre,method_scaling_large,method_from,method_cmat,method_copper}. 

However, with the enhancement of agent systems' capabilities and generalizability, the traditional dichotomy is inadequate for describing the diversity and hierarchy of memory in current agent systems. 
Building on the pioneering work by~\citet{survey_gaoling,survey_agentmemory}, we propose a more granular taxonomy that systematically categorizes agent memory along two fundamental dimensions (\S\ref{ssec:Memory Classification in Agents}). 
To more intuitively express our categorization, we have visualized specific examples in \autoref{fig:classification}.

\begin{figure*}[th]
    \centering
    \includegraphics[width=\linewidth]{figures/classification.pdf}
    \caption{Overview of the memory classification in agents. (a) Nature-based taxonomy that categorizes memory based on the type of information being encoded. (b) Scope-based classification that distinguishes memory according to how broadly it can be applied.}
    \label{fig:classification}
\end{figure*}

\subsection{Memory Classification in Cognitive Neuroscience}
\label{sssec:Memory Classification in Cognitive Neuroscience}
In cognitive neuroscience, memory can be divided into two main forms based on distinct temporal windows of information processing, thereby subserving their respective cognitive functions: 
(1) Short‑term memory (\S\ref{sssec:Short-term Memory}) enables the rapid, temporary maintenance and processing of incoming information, allowing an individual to respond promptly to the external environment
and (2) Long‑term memory (\S\ref{sssec:Long-term Memory}) is responsible for storing deeply processed experiences, which in turn can shape present cognition and guide future behavior.
In addition, we further discuss the distinction and interactions between them (\S\ref{sssec:Distinction and Interaction}).

\subsubsection{Short-term Memory}
\label{sssec:Short-term Memory}
Short-term memory refers to an information processing system that temporarily maintains and manipulates a small amount of information, with a time window generally not exceeding 15$\sim$20 seconds. 
Examples of using this memory include holding a phone number in mind for a few moments or keeping track of the last few sentences your conversation partner has just said. 
Due to limited cognitive resources~\cite{brain_class2}, the capacity of short-term memory is constrained such that it can only maintain 4$\sim$9 pieces of information simultaneously.
This ability varies substantially across individuals and influences human cognitive functions and behavioral performance, such as learning ability~\cite{brain_class_new_STM-1,brain_class_new_STM-2}, emotion regulation~\cite{brain_class_new_STM-3}, and creativity~\cite{brain_class_new_STM-4}.
When the amount of maintained information approaches this capacity limit, the brain dynamically reallocates its memory resources, prioritizing information that is more important or task-relevant and suppressing lower-priority representations~\cite{brain_class3}. 
If information is not transferred into long-term storage before this short-term time window closes, it is likely to be forgotten.

\subsubsection{Long-term Memory}
\label{sssec:Long-term Memory}
Long-term memory supports the storage and management of large amounts of information over extended periods. 
Its temporal span ranges from several minutes to many years or even decades. 
Examples include recalling a frequently used phone number or drawing on accumulated knowledge to offer original insights in a conversation.
Long-term memory provides an archive of past events and learned knowledge, which can be directly retrieved during interaction with the environment or used as a background context that shapes perception and decision-making~\cite{brain_class_LTM_intro_as_contex}. 
Unlike short-term memory, long-term memory is not characterized by a strict capacity limit. 
Instead, one of its key properties lies in the dynamic nature of information processing, including how it interacts with short-term memory (\S\ref{sssec:Distinction and Interaction}), how stored representations transform over time (\S\ref{sssec:Long-term Memory Storage}), and how information is managed (\S\ref{ssec:Memory Management in Cognitive Neuroscience}).

Based on the content of memories, long-term memory can be further divided into episodic memory and semantic memory. 
These two memories represent distinct types of information, with episodic memory encoding specific events and semantic memory storing abstract knowledge. 
This distinction is closely linked to significant differences in their underlying neural mechanisms. 
Furthermore, these memory systems do not operate in isolation within the brain but rather interact with each other and transform into one another over time.

\textbf{Episodic memory} 
refers to memory for specific events that an individual has personally experienced. 
Such memories typically include not only detailed information about the event itself, but also its temporal and spatial context—that is, when and where it occurred. 
Recalling an episodic memory is usually accompanied by a subjective sense of “mental time travel”~\cite{brain_class_LTM_e_s_mental_travel}, in which individuals feel as though they are transported back to the original situation, re-experiencing the surrounding environment and event details. 
For instance, remembering the drive to a newly opened cinema last week and the moment when someone made loud noises during the screening relies heavily on intact episodic memory.

\textbf{Semantic memory} 
refers to memory for learned factual knowledge, concepts, and rules. 
These memories are not tied to a specific time and place of acquisition, and their retrieval is not accompanied by a vivid re-experiencing of a particular past episode. 
For instance, knowing where a familiar building is located, or recalling the personality traits of an actor who played a given character in a film, depends primarily on semantic memory.

\subsubsection{Distinction and Interaction} 
\label{sssec:Distinction and Interaction}
From a functional perspective, short-term memory and long-term memory differ in the systems and timescales that support them. 
Short-term memory operates on newly encountered information and maintains it over seconds~\cite{brain_class1}, enabling rapid responses to the environment~\cite{brain_class_STM_LTM_distinct_1}. 
Long-term memory acts on information that has already been encoded, relying on slower learning processes to retain experiences and knowledge for years or even decades~\cite{brain_class_STM_LTM_distinct_2}.
More interesting than their differences are the interactions between them. 
In the classic multi-store model~\cite{brain_class_multi_memory_model}, short-term memory is described as a temporary “workspace” through which external information passes before entering long-term storage. 
Information that continues to be relevant is encoded into long-term memory, and when it is needed later, it is retrieved back into short-term memory to support ongoing processing. At the neural level, the two also interact. 
On the one hand, items that elicit stronger activity during short-term memory maintenance are more likely to be consolidated into durable long-term memories~\cite{brain_class4}, and higher working-memory load is associated with stronger hippocampal–neocortical coupling~\cite{brain_class5}. 
On the other hand, long-term memory provides priors and learned representational structures that shape how new information is encoded and maintained in short-term memory~\cite{brain_class6,brain_class7}. 
Thus, short-term and long-term memory are not isolated subsystems but mutually influential components of a memory network.

The differences between episodic and semantic memory extend beyond their content and subjective sense to their neural mechanism. 
For example, the individuals with hippocampal damage often have great difficulty reconstructing the specific scenes of past events, yet their semantic memory is relatively preserved~\cite{brain_class8}. 
Evidence from imaging studies indicates that these two types of memory rely on partially dissociable brain systems. 
Episodic memory depends strongly on the hippocampus, whereas semantic memory is supported primarily by the neocortex~\cite{brain_class_LTM_e_s_diff_1,brain_class_LTM_e_s_diff_2,brain_class_LTM_e_s_diff_3}.
In real life, episodic memory and semantic memory are usually intertwined and interact with each other. 
Repeatedly experiencing similar events allows the brain to extract stable structures and rules~\cite{brain_LTM_storage15}, gradually forming more abstract semantic knowledge and achieving a transformation from episodic to semantic form. 
In turn, when we recall a specific episode, existing semantic knowledge can serve as a prior context to guide its reconstruction, filling gaps and sometimes distorting details~\cite{brain_class9}. 
Thus, although episodic and semantic memory are conceptually distinct, they continuously influence and reshape each other over time.

\subsection{Memory Classification in Agents}
\label{ssec:Memory Classification in Agents}
A coherent taxonomy of memory is essential for systematically understanding and designing memory mechanisms in agent systems. 
In this section, we introduce two complementary classification frameworks: 
(1) Nature-based taxonomy (\S\ref{sssec:Nature-Based classification}) that distinguishes memory according to the type of information it encodes, 
and (2) Scope-based taxonomy (\S\ref{sssec:Scope-based classification}) that characterizes memory by its boundaries of applicability across tasks or sessions.

\subsubsection{Nature-based Classification}
\label{sssec:Nature-Based classification}
We observe that the nature of memory in agent systems closely parallels that found in cognitive neuroscience research. 
This nature is fundamentally determined by the type of information memory provides to subsequent reasoning processes.
Building upon this correspondence, we adopt the classical taxonomy from cognitive neuroscience and categorize agent memory into two types, namely episodic and semantic memory, as shown in \autoref{fig:classification} (a).

\textbf{Episodic memory} refers to the agent's experiential memory that stores sequential interaction trajectories and contextual information. 
This memory type is tool-augmented, maintaining detailed logs of what tasks were attempted, which tools were invoked, and what solution pathways were followed. 
It captures the procedural history of the agent's problem-solving processes, including intermediate steps, tool call sequences, and decision branches, enabling it to learn from past execution patterns and optimize future task completion strategies.

\textbf{Semantic memory} functions as the agent's knowledge repository without tool dependencies. 
It stores factual information, concepts, rules, and general knowledge~\cite{method_reasoningbank}. 
This memory type provides the foundational understanding necessary for reasoning and inference, containing declarative knowledge such as definitions, relationships between concepts, and general principles that guide the agent's behavior.

\textbf{Distinction.} The fundamental distinction between the two lies in the nature of the content they aim to convey. 
Episodic memory aims to convey experiential information, recording procedural knowledge of “how to do things”, while semantic memory aims to convey conceptual information, storing declarative knowledge of “what things are”.
More importantly, this classification not only reflects the organizational structure of memory content but also reveals that agents need to coordinate two complementary cognitive strategies, experience- and knowledge-driven, when handling complex tasks.

\subsubsection{Scope-based Classification}
\label{sssec:Scope-based classification}
Memory in agents can be classified based on its scope of applicability. 
This scope-based classification determines whether memory is confined to a single task or session, or extends across multiple tasks or sessions.
Accordingly, it can be divided into two categories, namely inside-trail and cross-trail memory, as shown in \autoref{fig:classification} (b).

\textbf{Inside-trail memory} 
is confined to a single trajectory execution. 
It stores context-specific information such as intermediate steps, temporary variables, and task-relevant observations that are only valid within the current episode~\cite{method_mem1,method_deepagent,method_agentfold,method_foldgrpo,method_resum,method_memory_as_action,dy_memory_mem,method_memtool}. 
For instance, \citet{method_mem1} proposed maintaining temporary state information for each reasoning step during an agent's problem-solving process, which was then propagated to subsequent steps.
This memory is typically cleared or reset when the episode ends.

\textbf{Cross-trail memory} 
persists across multiple trajectory executions, enabling agents to accumulate knowledge and experience over time. 
It stores generalizable patterns, learned strategies, and reusable knowledge that can inform future episodes~\cite{method_reasoningbank,method_legomem,method_playbook,method_evolver}. 
This persistent memory facilitates continual learning and adaptation.
For example, \citet{method_reasoningbank} proposed long-term storage of historical successful and failed trajectories in a memory repository, where each memory entry was organized into a structured format consisting of title, description, and content, serving as an experiential knowledge base for subsequent task processing.

\textbf{Distinction.} The key distinction lies in temporal range and reusability. 
Inside-trail memory is transient and trajectory-specific, providing essential working space for complex reasoning and multi-step problem solving within episodes. 
In contrast, cross-trail memory is persistent and generalizable, transforming historical trajectories into strategic knowledge to help agents systematically improve their performance over time.

%% file: 5_Storage.tex
\section{Memory Storage}
\label{sec:Storage}
Memory storage is a core function of cognitive systems, determining how information is retained and organized. 
This section systematically examines two key dimensions of memory storage, namely storage location and storage format, from the perspectives of both cognitive neuroscience (\S\ref{ssec:Memory Storage in Cognitive Neuroscience}) and agents (\S\ref{ssec:Memory Storage in Agents}). 
Comparing the similarities and differences between these two types of systems provides deeper insights into the fundamental mechanisms of memory storage.

\subsection{Memory Storage in Cognitive Neuroscience}
\label{ssec:Memory Storage in Cognitive Neuroscience}
Memory storage in the human brain is not a unitary process but rather a dynamic interplay between multiple systems operating at different timescales. 
Understanding how the brain retains information requires distinguishing between short-term memory, which holds information temporarily for immediate use, and long-term memory, which preserves experiences and knowledge over extended periods.
Although these two systems differ in their temporal scope and neural substrates, they are interconnected with respect to memory storage.
Short-term representations can be consolidated into long-term traces, and long-term memories can be reactivated into short-term working states. 
In the following sections, we examine the neural architecture and representational formats underlying short-term memory (\S\ref{ssec:Short-term Memory Storage}) and long-term memory (\S\ref{sssec:Long-term Memory Storage}), drawing on converging evidence from neuroimaging, electrophysiology, and lesion studies in humans.

\subsubsection{Short-term Memory Storage}
\label{ssec:Short-term Memory Storage}
Short-term memory, as a temporary information storage system, primarily processes a small amount of information within a limited time window. 
To deeply understand how short-term memory supports flexible cognition, however, it is necessary to consider where in the brain short-term information is stored and in what form it is represented and maintained. 
A large body of neuroimaging and electrophysiological evidence has begun to outline the neural basis of short-term memory. 
On the one hand, it appears to rely on a distributed storage scheme spanning sensory cortices and frontoparietal network (i.e, sensory–frontoparietal network). 
On the other hand, at the cellular and circuit levels, information can be held via mechanisms such as persistent neural firing and activity-silent synaptic connection states. 
In this section, we briefly review recent work on short-term memory along two dimensions, namely its storage location and its underlying storage format.

\textbf{Storage Location.} 
Human neuroimaging studies consistently show that short-term memory is supported by a distributed brain system. 
After the stimulus disappears, information about the remembered items remains retained in the sensory cortices, posterior parietal cortex, and prefrontal regions. 
During the maintenance period, these areas display persistent activity or decodable patterns that match the memory content. 
This means short-term memory is not stored in one single region, instead, it relies on sensory–frontoparietal network.
Specifically, perceptual cortices tend to retain fine-grained sensory details, while the prefrontal cortex mainly regulates this storage process by setting priorities, allocating limited memory resources across items~\cite{brain_STM_storage1}, and recoding information into formats that better meet upcoming behavioral demands~\cite{brain_STM_storage2}.
In addition, frontoparietal network also supports cross-modal representations, allowing information from different sensory channels to be linked and manipulated in a shared representational space~\cite{brain_STM_storage3}.

\textbf{Storage Format.} 
At the cellular and circuit levels, short-term memory can be maintained by more than one format, which include: Persistent activity and Synaptic connection weights.

\begin{itemize}
	\item \textbf{Persistent activity} refers to a classic proposal in which neurons that encode the memory maintain a high level of firing activity even after the external stimuli are removed. 
    Single-neuron recordings in epilepsy patients provide direct human evidence for this idea, showing that neurons in the medial frontal and medial temporal lobes remain active after stimulus offset and their firing tracks the remembered content~\cite{brain_STM_storage4}.
    However, persistent activity is energy-demanding and may be vulnerable to interference.
	\item \textbf{Synaptic connection weights} refer to an alternative format for memory storage, where information can be temporarily maintained without sustained neural activity. 
    In this framework, population firing of neurons can drop back to baseline while the memory enters an “activity-silent” state that standard recordings cannot easily decode~\cite{brain_STM_storage5}. 
\end{itemize}

A more widely accepted view today is that both of the formats coexist, with the brain switching between them. 
High-priority items within the focus of attention are often kept in an active, persistently firing state, while items outside attention but still potentially useful can be stored silently and reactivated when needed~\cite{brain_STM_storage6}. 
To support this dual-mechanism hypothesis, \citet{brain_STM_storage7} showed that electroencephalography (EEG) can decode memory content during maintenance, but not during the inter-trial interval, demonstrating how the representation becomes latent and later re-emerges.

\begin{figure*}[t]
    \centering
    \includegraphics[width=\linewidth]{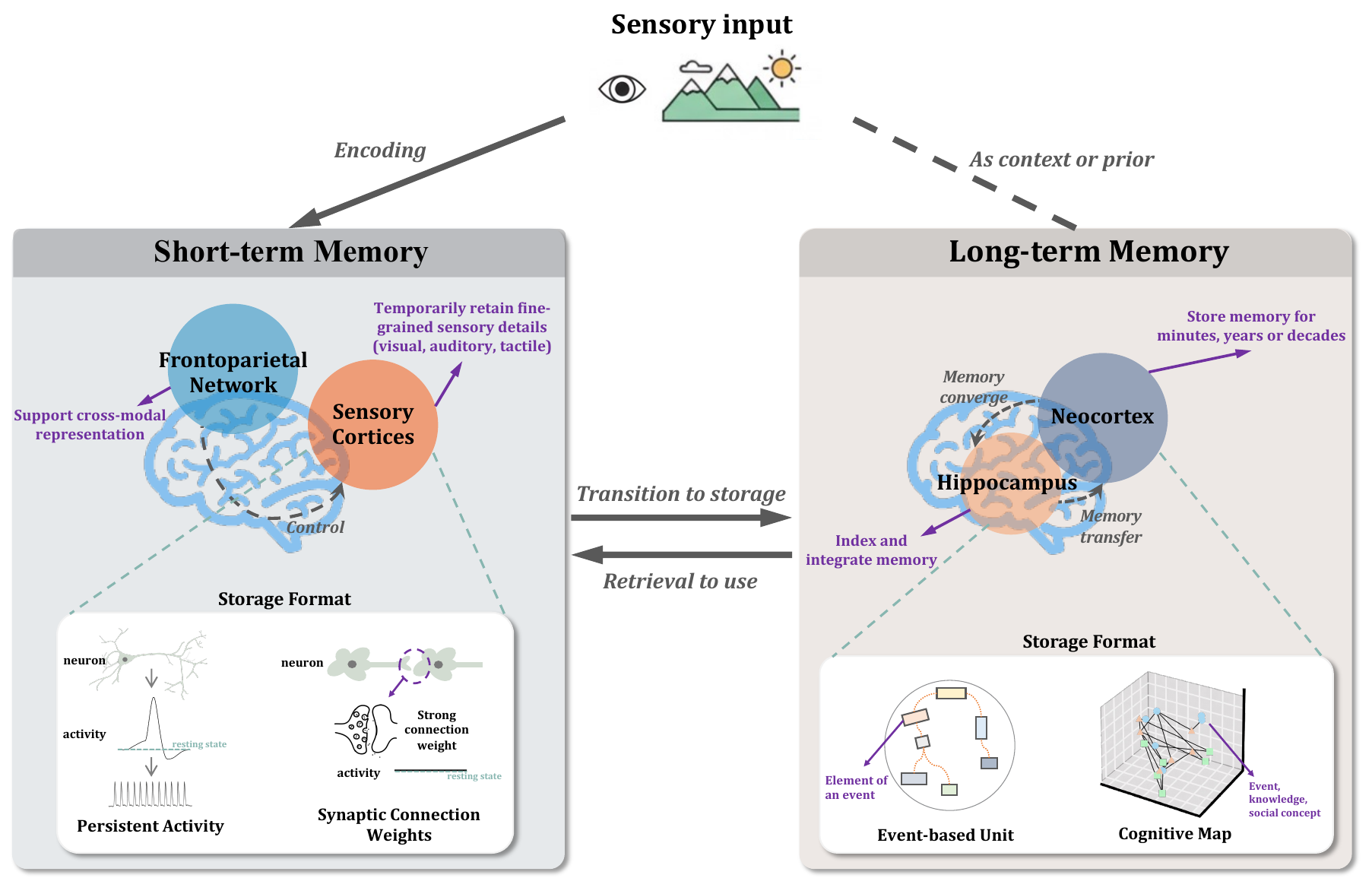}
    \caption{Overview of memory storage mechanisms in cognitive neuroscience, including storage locations and storage formats of short- and long-term memory.}
    \label{fig:storage_long_short}
\end{figure*}

\subsubsection{Long-term Memory Storage}
\label{sssec:Long-term Memory Storage}
Compared with the temporary maintenance of information by short-term memory, the characteristic of long-term memory that it can preserve information for a long time implies that it has a relatively more complex neural mechanism. 
Rather than preserving experiences in their original details, the brain gradually transforms and reorganizes them across hippocampal–neocortical systems, giving rise to more structured and abstract representations. 
In this section, we will introduce where in the brain long-term memories are stored, and in what structural form they are represented.
Specifically, we first summarize evidence on the division of labor between the hippocampus and neocortex in long-term memory storage, and then discuss how experiences are organized into structured units, such as event-based unit and cognitive map.

\textbf{Storage Location.}
Long-term memory storage depends on the coordinated activity of two key brain regions, the hippocampus and the neocortex.
When new information is acquired, it is initially encoded and maintained in distributed neocortical regions, whose signals then converge in the hippocampus for integrated processing. 
Rather than serving as a storage warehouse, the hippocampus functions as an index that reactivates these distributed memory traces.
Consequently, newly formed memories depend heavily on hippocampal support, but through systems consolidation, they gradually become sustained by neocortical networks~\cite{brain_LTM_storage1}.
Multiple findings support this view. 
For example, \citet{brain_LTM_storage2} demonstrated that hippocampal activity during encoding predicts how precisely details will be stored, but later retrieval success depended more on prefrontal activity and the strength of neocortical representations. 
In another study, \citet{brain_LTM_storage3} found that rehearsal strengthened memories mainly in neocortical and hippocampal–neocortical interaction regions, rather than in hippocampal activity alone. 
Moreover, when people experience events with overlapping contexts, the hippocampus and relevant neocortical regions show coordinated reactivation during rest. 
This replay helps write cross-event structures into medial prefrontal cortex (mPFC) representations~\cite{brain_LTM_storage4}.These cross-event structures can then serve as contextual knowledge or priors that shape the encoding of information in short-term memory~\cite{brain_class6,brain_class7}.

\textbf{Storage Format.}
It is an obvious fact that memories cannot be preserved in their original details. 
During consolidation, the hippocampus integrates information and develops more abstract, structured representations, allowing memories to be stored in complex forms rather than as raw sensory data.
Here we focus on two such forms: event-based units and cognitive maps.

\begin{itemize}
	\item \textbf{Event-based unit} refers to the discrete memory representations that result from the brain's segmentation of continuous, rich everyday experience into distinct episodes, allowing us to remember life as a series of separate events rather than an unbroken stream. 
    An fMRI study showed strong hippocampal responses at boundaries between experimentally defined events, suggesting that the brain automatically detects event transitions~\cite{brain_LTM_storage5}.
    However, this segmentation does not produce entirely isolated memories. 
    Even when plots, scenes, or locations change, events that occur close in time are often recalled as a coherent stream. 
    This continuity is facilitated by hippocampal-neocortical replay at the end of an event, which reinstates what just happened and helps integrate information across boundaries~\cite{brain_LTM_storage6}.
    Beyond temporal integration, a complete event also appears to have a holistic representation. 
    Although its constituent elements are distributed across neocortical areas, the hippocampus can bind them into a unified memory trace~\cite{brain_LTM_storage7}.
    Consistent with this, human hippocampus contains event-specific neurons that fire strongly during encoding and retrieval of particular episodes. 
    Their activity reflects the event as a whole rather than isolated features~\cite{brain_LTM_storage8}.
	\item \textbf{Cognitive map} refers to the mental representations built by the hippocampus–entorhinal circuit, which not only tracks locations and routes during physical navigation but also serves a broader function in organizing abstract knowledge, concepts, and experiences. 
    Within such maps, past experiences are represented as points in a multidimensional space.
    The construction of cognitive maps involves several key processes. 
    First, the brain establishes an internal coordinate system for abstract concepts that remarkably resembles the one it uses for physical space.
    For instance, during memory for two-dimensional conceptual spaces, entorhinal cortex and ventromedial prefrontal cortex (vmPFC) show hexagonally symmetric signals, similar to grid-cell coding in navigation~\cite{brain_LTM_storage9}. 
    Related coordinate-like representations have also been found for social relationships~\cite{brain_LTM_storage10}. 
    Second, the hippocampus–entorhinal system can encode distances between knowledge units. 
    Neural signal strength reflects relational or path-like distance in the cognitive space~\cite{brain_LTM_storage11}. 
    This principle extends to event memories, which can be arranged into temporal maps where distances correspond to transition probabilities over time~\cite{brain_LTM_storage12}.   
    Third, through consolidation, these maps are gradually transferred to and maintained in neocortical regions. 
    It has been recorded that medial prefrontal cortex stores learned abstract relations~\cite{brain_LTM_storage13}, while other frontal and cingulate regions encode social and functional attributes that organize long-term knowledge~\cite{brain_LTM_storage14}.
\end{itemize}

\subsection{Memory Storage in Agents}
\label{ssec:Memory Storage in Agents}
Unlike the human brain with its complex structures and rich biological information, memory storage in agents is fundamentally based on natural language symbols. Therefore, rather than discussing storage by memory type, we directly introduce two key dimensions of the storage: 
(1) Storage location (\S\ref{sssec:Storage Location}) 
and (2) Storage format (\S\ref{sssec:Storage Format}).

\subsubsection{Storage Location}
\label{sssec:Storage Location}

From the perspective of storage mechanisms, memory in agents can be organized in two primary ways. 
The context window serves as a container that dynamically folds and updates information within a single reasoning trajectory. 
In contrast, memory banks function as external repositories that persistently store accumulated information. 

\textbf{Context Window.}
The context window can serve as a container for memory storage.
Information placed within the context, including user inputs, tool invocation result, and intermediate generation steps, can be transformed into inside-trail memory that the agent directly accesses during reasoning. 
This process is defined as memory folding~\cite{method_mem1,method_deepagent,method_agentfold,method_foldgrpo,method_resum,method_memory_as_action,dy_memory_mem,method_memtool}, which is triggered during the reasoning process, enabling the agent to dynamically update its state of knowledge within its trajectory.
For example, \citet{method_resum} proposed periodically calling a summarization tool to compress accumulated interaction history into a structured summary, allowing the agent to resume reasoning from the compressed state and thus mitigating truncation issues caused by context window limitation.
Building upon this concept, some works~\cite{method_memory_as_action,method_deepagent} have further proposed active folding strategies. 
For example, \citet{method_deepagent} proposed to dynamically determine whether to initiate memory folding during reasoning by generating a special trigger token, thereby providing greater flexibility and adaptability for memory management.

\textbf{Memory Bank.}
Beyond context-based storage, agent systems commonly incorporate external memory modules as dedicated repositories.
Unlike the context window, such memory module~\cite{method_agentkb,method_rmm,method_aios,method_reasoningbank,method_reactree,method_memory_sharing} possess theoretically unbounded capacity and can persistently retain the agent's cross-trail memory, which includes experiential data and domain knowledge accumulated across multiple sessions and tasks.
This type of memory storage mode is typically triggered at the end of a trajectory, granting memory reusability, ensuring that previously acquired useful information remains accessible and retrievable even after conversation termination or system restart.
Specifically, some memory banks are designed to persistently store user preferences, knowledge, and conversation history. 
For example, \citet{method_rmm} introduced a framework that organize the memory bank in a topic-based manner, enabling cross-session tracking of users’ health conditions, allergies, and personal preferences for continuous personalized service.
Other memory banks focus on distilling reusable strategies and reasoning patterns from agents’ historical interactions. 
For instance, \citet{method_reasoningbank} suggest to extract high-level reasoning strategies from both successful and failed experiences, storing abstracted decision principles rather than raw trajectories.
Building upon these approaches, cross-agent shared memory banks transcend the boundaries of individual agents, enabling experience sharing across different frameworks. 
\citet{method_agentkb} propose plug-and-play knowledge sharing by abstracting execution trajectories from various agent systems into unified structured experience units, accessible through lightweight APIs. 
This design addresses the isolation problem in current agent systems, allowing solutions discovered in one framework to be directly reused by others and avoiding redundant trial and error.

\subsubsection{Storage Format}
\label{sssec:Storage Format}
Memory representations in agents are typically divided into four categories, including text stored in conventional natural language, graphs that emphasize preserving structured relationships, parametric memory internalized in model weights, and latent representations in vector form.

\textbf{Text.}
Natural language text is the most common storage format for agent memory. Memories are stored as either raw text or summarized text, encompassing both experiences and information~\cite{method_ace,method_hiagent,method_mms,method_apc,method_eapo,method_maple,method_avatar}. 
This format offers high interpretability, ease of manipulation, and direct compatibility with language model architectures. 
Although stored in textual form, conversational or task-related processes are typically not preserved verbatim but undergo a certain degree of abstraction and summarization. 
For example, \citet{method_hiagent} proposed decomposing tasks into multiple sub-goals, with each sub-goal completion triggering a consolidation of multiple steps into a summary that is then stored in the memory container.
In contrast, \citet{method_ace} argued that excessive abstraction and compression of memory information causes agents to suffer from brevity bias, leading to degraded performance in specialized domains. 
They therefore proposed the playbook memory storage mechanism, which minimizes memory compression and preserves detailed information to the greatest extent possible.
Overall, text-format memory facilitates retrieval and combination while fully leveraging the language processing capabilities of large language models for reasoning.

\textbf{Graph.}
Graph-based memory storage organizes memory into a structured network composed of entities and relationships~\cite{method_fes,method_sgmem,method_flexibly,method_from_isolated,method_arigraph,method_towards_lifelong,method_crafting}. 
In this format, experiences or information are decomposed into nodes (e.g., concepts, objects, events, or steps) and edges (e.g., relationships). 
The graph structure excels at supporting complex reasoning tasks by allowing the system to traverse edges between nodes and concatenate multiple episodes into the most effective memory as experience.
For instance, \citet{method_mem0} constructed a relational graph to explicitly model temporal and logical relationships between entities, demonstrating strong performance in tasks requiring multi-step reasoning by effectively tracking event sequences and character interactions. 
\citet{method_gmemory} employed a three-tier graph hierarchy to manage the lengthy interaction history of multi-agent systems, retrieving high-level, generalizable insights and fine-grained collaboration trajectories through bi-directional memory traversal.
Furthermore, the graph structure inherently supports relationship extraction and pattern discovery, enabling the identification of implicit connections between nodes and thereby assisting complex logical information queries. 
For example, \citet{method_dsmart} constructed the dialogue history into a knowledge graph and executes multi-step graph traversal searches, making the reasoning process traceable and faithful to the dialogic facts. 
\citet{agent_memory_see} proposed an entity-centric multimodal graph to store memories of faces, voices, and text, connecting nodes across different modalities through identity equivalence detection to maintain long-term consistency.
Overall, graph-structured memory enhances reasoning performance and consistency in complex scenarios through its inherent relational representation capabilities.

\textbf{Parameters.}
Parametric storage embeds memories directly within model weights, integrating experience and knowledge into the neural architecture~\cite{method_pretraining_context}. 
This approach solidifies information through persistent modifications of connection strengths, thereby emulating synaptic plasticity in biological brains. 
Its primary advantages include exceptional access efficiency and deep knowledge integration, as memory is automatically activated during forward propagation, which eliminates retrieval latency and cross-modal alignment costs.
This internalization of experience into neural weights is primarily driven by imitation and reinforcement learning. In terms of imitation learning, agents utilize supervised fine-tuning to shape initial parametric representations. 
For instance, \citet{method_memverse} employed distillation to solidify long-term knowledge within weights, while \citet{method_seagent} distilled expert trajectories to transform operational skills into parametric instincts. Conversely, reinforcement learning achieves experience-driven learning. Under this paradigm, agents treat sampled trajectories as episodic memories. Specifically, \citet{method_digirl, method_webrl, method_webagenrr1} successfully internalized explored strategies into parameters by extracting performance advantages from these samples. 
Furthermore, \citet{method_early_experience} introduced an intermediate paradigm that enables agents to learn from self-interaction without explicit rewards by using future states of exploratory actions as supervision signals to internalize experience.
In summary, by internalizing experiences into model weights, parametric storage eliminates retrieval latency and enhances decision stability, thereby enabling rapid and consistent agent responses.

\textbf{Latent Representation.}
Beyond token-space memory structures that explicitly represent information and parametric memory, several studies~\cite{method_cam, method_r3mem} have explored latent memory, which stores information as high-dimensional vectors in embedding space. 
Compared to token-level memory, latent memory offers several key advantages:
(1) Efficient compression: high-dimensional continuous vectors can encode information more compactly than discrete tokens, thereby reducing storage and computational overhead~\cite{cog_memory_memllm,method_rf}. 
(2) End-to-end trainability: as continuous representations, latent memories can directly participate in gradient-based optimization, enabling updates and refinements during training~\cite{method_em2,method_icae}.
(3) Alignment with human cognition: as noted in~\cite{method_nature,method_coconut}, human reasoning relies on integrated representations that transcend discrete symbols, latent space representations effectively embody this principle.
Representative methods include MemoryLLM~\cite{cog_memory_memllm} and M+~\cite{method_m+}, which allocated a fixed set of latent vectors as “internal memory” at each layer of LLMs. 
As the context evolves, these latent memories are iteratively retrieved and updated, then concatenated with the hidden states at each layer and integrated into the model’s forward computation, thereby enabling the retention and utilization of information from long documents.
Building upon this, \citet{method_memgen} further extended this approach to the domain of agent memory. 
It fed the hidden state of the agent's current action into a trigger module to determine whether latent memory needs to be invoked. 
When triggered, the agent leveraged the current action's hidden state to retrieve relevant latent memories from Weaver's parameters, dynamically steering the model's subsequent actions and enabling continuous self-evolving behavior.

%% file: 6_Management.tex
\section{Memory Management}
\label{sec:Management}

Memory management represents a dynamic regulatory framework governing the entire information lifecycle, ensuring that intelligent systems can adaptively filter noise and preserve pivotal patterns within changing environments. 
Whether through the consolidation of labile traces in biological brains or the regulation of extensive storage in artificial systems, effective management mechanisms are central to transforming raw perception into structured knowledge. 
This chapter provides a cross-disciplinary analysis of memory management, beginning with its neuroscientific foundations (\S\ref{ssec:Memory Management in Cognitive Neuroscience}) and proceeding to the systematic implementation of information-flow management in autonomous agents (\S\ref{ssec:Memory Management in Agents}).

\subsection{Memory Management in Cognitive Neuroscience}
\label{ssec:Memory Management in Cognitive Neuroscience}

\begin{figure*}[th]
    \centering
    \includegraphics[width=\linewidth]{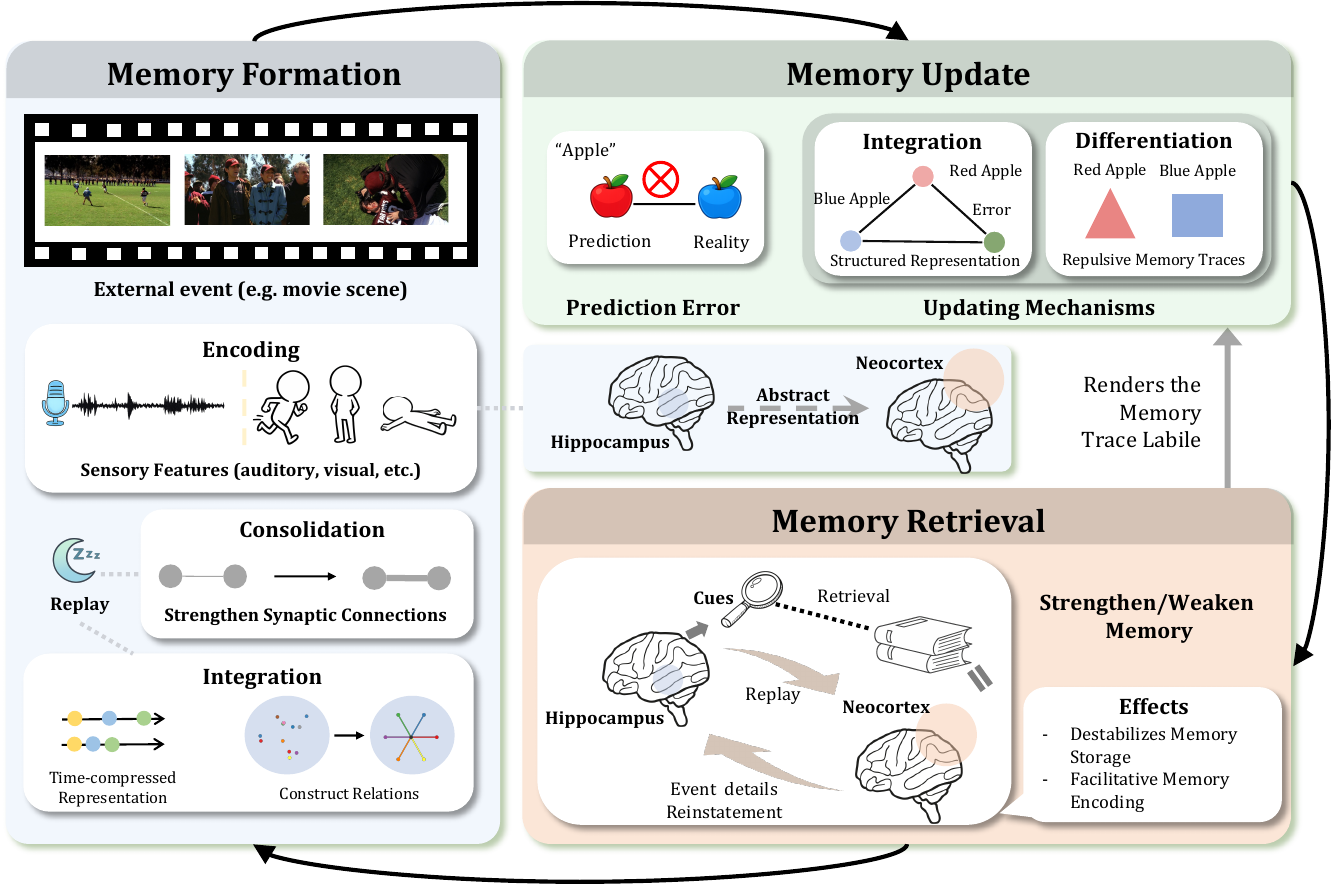}
    \caption{Overview of memory management in cognitive neuroscience. The framework illustrates a dynamic cycle of information processing including memory formation, updating, and retrieval, through which long-term memory supports flexible adaptation to the external environment.}
    \label{fig:management_neural}
\end{figure*}

In the field of cognitive neuroscience, long-term memory exhibits more sophisticated dynamic processes over extended temporal scales compared to short-term memory which involves only the transient maintenance of information within a scale of seconds. 
Memory management discussed in this section focuses primarily on long-term memory. 
As illustrated in \autoref{fig:management_neural}, management processes, which consist of formation, updating, and retrieval, interact reciprocally with storage mechanisms to constitute a dynamic cycle of information processing. 
As the functional inception of this cycle, memory formation transcends rudimentary encoding as it is intrinsically intertwined with consolidation and integration, which collectively shape the content and structural architecture of stored information. 
Conversely, updating and retrieval act as functional drivers that re-initiate the formative cycle by triggering the re-encoding and reconsolidation of existing memory traces. 
This section will first delineate how memory representations are instantiated and gradually stabilized through the lens of memory formation (\S\ref{sssec：Memory Formation}), and then address the regulatory roles of updating (\S\ref{sssec:Memory Updating}) and retrieval (\S\ref{sssec:Memory Retrieval}) within the memory system.

\subsubsection{Memory Formation}
\label{sssec：Memory Formation}
Long-term memories do not emerge as fully instantiated entities at the moment of experience.
Instead, they develop through a sequential process of encoding, consolidation, and integration. 
Initially, ongoing events are encoded as specific neural activity patterns across hippocampal and neocortical circuits. 
Subsequently, during periods of wakeful rest and sleep, the hippocampus replays these events to reactivate the brain's initial activity patterns, thereby stabilizing memory representations through system consolidation. 
Over time, memories undergo temporal compression and cross-episodic linkage, ultimately being stored in the neocortex as abstract structural representations.
In the following, we sequentially delineate these three processing stages that underpin memory formation.

\textbf{Encoding.}
During an experience, neuronal ensembles in the hippocampus and neocortex with elevated intrinsic excitability are selectively recruited, transforming event information into memory codes through excitatory population firing. 
In this process, the hippocampus binds the distributed sensory features within the neocortex into a unified representation and selectively modulates its interaction with sensory cortices to amplify representations of high future utility~\cite{brain_memory_manage1,brain_memory_manage2}. 
This encoding phase engages long-term synaptic plasticity.
Following the Hebb principle~\cite{brain_memory_manage3,brain_memory_manage4}, synchronized pre- and post-synaptic activity strengthens connections (LTP), whereas desynchronized activity leads to weakening (LTD). 
Such persistent modifications in synaptic efficacy underpin the biological foundation of long-term memory storage.

\textbf{Consolidation.}
System consolidation is essential for newly formed memories to achieve long-term stability and retrieval precision~\cite{brain_memory_manage5}. 
During offline states such as wakeful rest or sleep, the hippocampus spontaneously replays recent experiences~\cite{brain_memory_manage6,brain_memory_manage7}, triggering the reactivation of neural patterns within hippocampal-neocortical circuits~\cite{brain_memory_manage8, brain_memory_manage9}.
Throughout these replay events, transient bursts of hippocampal activity synchronize with rhythmic neocortical activity, enabling the two regions to “fire together” with high temporal precision~\cite{brain_memory_manage10}. 
Within this inter-regional dialogue, the hippocampus projects discrete recent experiences to the neocortex, utilizing existing neocortical knowledge structures as a scaffold to reorganize and align this new information for integration into the long-term storage framework~\cite{brain_memory_manage11,brain_memory_manage12}.

\textbf{Integration.}
Integration serves as the culmination of the memory formation cycle, functioning to assimilate consolidated, discrete memories into the brain’s broader knowledge structures. 
This process is intrinsically linked to encoding and consolidation; while encoding captures information and consolidation preserves it, integration transforms these stabilized traces into organized, relational knowledge.
Facilitated by hippocampal replay, memories undergo qualitative transformations where event sequences are temporally compressed~\cite{brain_memory_manage13} and overlapping elements across distinct episodes are cross-linked~\cite{brain_memory_manage14}. 
During this stage, the hippocampus and the medial prefrontal cortex (mPFC) collaborate to identify and construct logical associations between events. 
Integrated information is then gradually redistributed to specific neocortical regions, supporting more enduring and abstract forms of storage.

\subsubsection{Memory Updating}
\label{sssec:Memory Updating}
Once memory representations are established and stabilized through the aforementioned formation processes, they do not remain static but are subject to continuous dynamic evolution. 
To adapt to an ever-changing environment, humans must constantly refine and update their stored knowledge. 
During this process, individuals generate expectations about upcoming events based on prior memories. 
When a discrepancy arises between reality and these expectations, the resulting prediction error serves as a pivotal signal that triggers memory updating, driving the brain to reshape previously stabilized representations~\cite{brain_memory_update1}.

Memory updating transcends the mere replacement of outdated information, instead relying on sophisticated mechanisms such as differentiation and integration. 
On one hand, updating can be achieved through differentiation, where the hippocampus generates distinct and mutually repulsive neural representations for highly similar events. 
This process preserves mnemonic specificity while mitigating interference between old and new information~\cite{brain_memory_update2,brain_memory_manage25}. 
On the other hand, updating frequently involves integrative mechanisms. According to the integrative updating theory proposed by \cite{brain_memory_update3}, individuals organize new information, retrieved prior memories, and the associated prediction errors into a unified, structured representation. 
A salient example is the correction of misinformation, where individuals can successfully retain the factual truth while simultaneously maintaining a memory of the original fake news and its discrepancy from the facts~\cite{brain_memory_update4}.

\subsubsection{Memory Retrieval}
\label{sssec:Memory Retrieval}
Although memories are encoded and stored through the aforementioned processes, they typically persist in a latent state, influencing current cognitive processing and behavior only upon successful retrieval. 
This process is generally initiated by external cues or contextual information. For instance, encountering a stimulus that resembles a constituent element of a past experience, or re-entering a familiar environment, acts as a functional trigger. 
Through the mechanism of hippocampal pattern completion, these partial cues facilitate the reconstruction and retrieval of the entire episodic representation~\cite{brain_memory_manage16}.

\textbf{The Neural Basis of Retrieval.}
At the neural level, retrieval involves rapid and temporally compressed replay of episodic content stored within the hippocampus. 
The sequence of these replays is flexible and adapts to specific tasks. For instance, following the acquisition of sequential items, tasks requiring the recall of items preceding a cue trigger reverse hippocampal replay, whereas tasks focusing on subsequent items elicit forward replay matching the original experience~\cite{brain_memory_manage17}. 
As replay unfolds, the associative representations between events and their contexts are typically reconstructed in the hippocampus prior to the emergence of specific details. 
Subsequently, hippocampal burst activity guides the neocortical reinstatement of fine-grained episodic content~\cite{brain_memory_manage18,brain_memory_manage19}, even including representations at the sensory feature level~\cite{brain_memory_manage20}. 
As mnemonic content expands temporally, various neocortical regions are sequentially engaged to support the dynamic reconstruction of naturalistic autobiographical memories~\cite{brain_memory_manage21}.

\textbf{The Impact of Retrieval on Existing Memory Storage.}
Retrieval is not a passive readout of stored information but a transformative process that modifies memory traces. According to reconsolidation theory, the retrieval of a memory renders the trace labile within a temporary window, allowing it to be strengthened, weakened, or updated during subsequent restabilization. 
Repeated retrieval of an event causes the co-encoded contextual background to be repeatedly reactivated, which progressively evolves into a potent retrieval cue, thereby enhancing future retrieval success and memory durability~\cite{brain_memory_manage22}. 
This facilitation extends to other related memories sharing the same context through similar reactivation processes~\cite{brain_memory_manage23}. 
Furthermore, retrieval-related enhancement promotes the integration of new and old information. 
When these memories are interrelated, retrieval practice of new information induces the alternating reinstatement of both encoding contexts~\cite{brain_memory_manage24}. 
During this process, elevated activity in the mPFC facilitates the differentiation and integration of these representations, leading to more robust storage of new memories and a biasing or partial overwriting of older traces~\cite{brain_memory_manage25}. 
However, this plasticity can have maladaptive consequences, such as the reconsolidation of erroneous details introduced during retrieval, a process governed by the functional balance between conflict-processing networks and sensory integration systems~\cite{brain_memory_manage26}.

\textbf{The Facilitative Effect of Retrieval on Subsequent Encoding.}
Beyond modifying existing traces, retrieval serves as a functional primer that enhances the encoding efficiency of subsequent new information. 
Mechanistically, retrieval engages frontal-hippocampal networks to optimize the brain's cognitive state, thereby directly improving the re-encoding and integration of incoming information~\cite{brain_memory_manage22,brain_memory_manage27}. 
Additionally, retrieval practice activates reward- and prediction-related regions, including the ventral striatum, insula, and midbrain. 
This activity shifts the brain into a predictive learning mode, heightening both motivation and sensitivity to novel experiences~\cite{brain_memory_manage28}. 
Notably, this facilitative effect is a double-edged sword: while it improves general learning efficiency, it may also expedite the internalization of misleading information if the subsequent input is inaccurate.

\subsection{Memory Management in Agents}
\label{ssec:Memory Management in Agents}

\begin{figure*}[th]
    \centering
    \includegraphics[width=\linewidth]{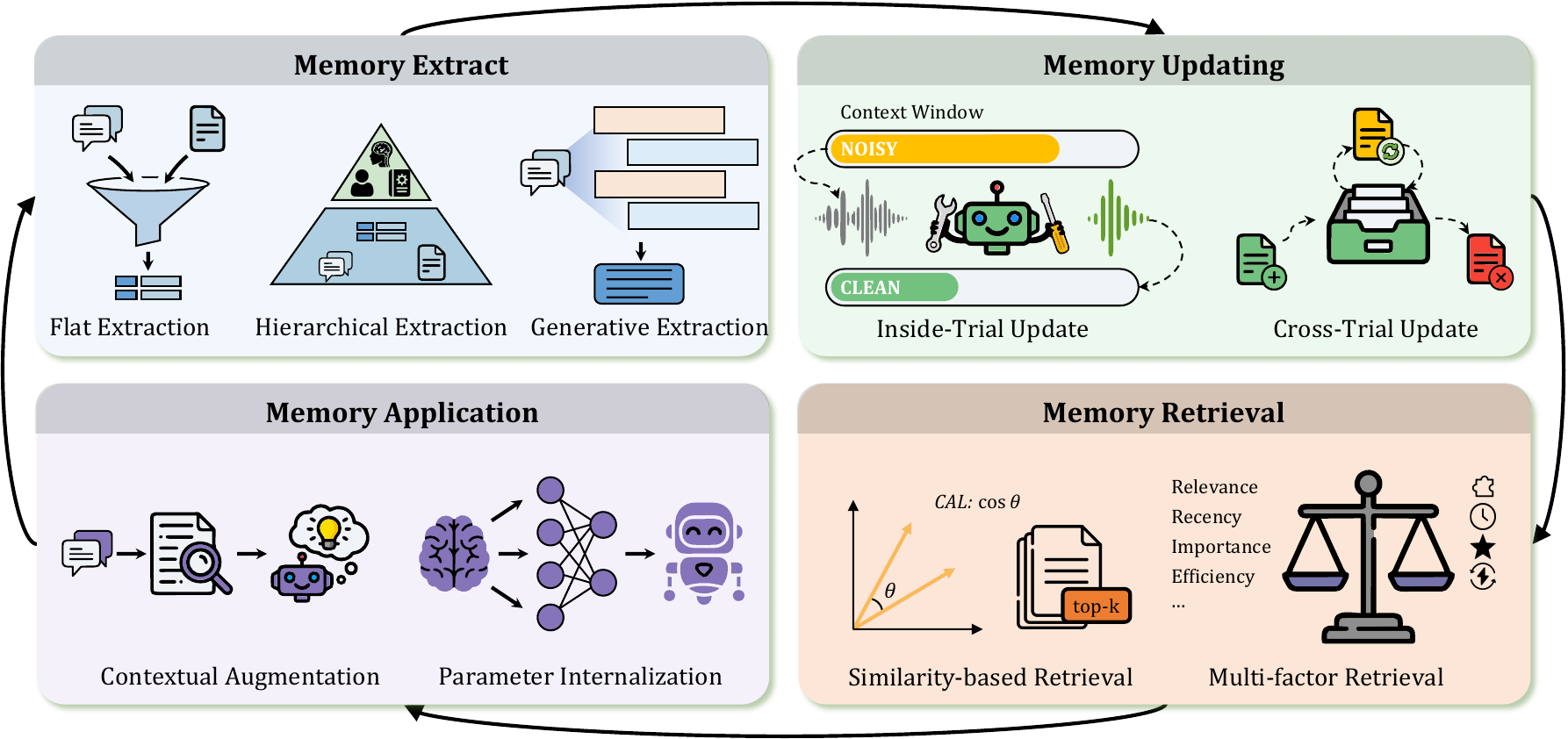}
    \caption{Overview of memory management in agents. The framework forms a closed-loop pipeline consisting of memory extraction, updating, retrieval, and utilization, enabling persistent experience regulation and long-range reasoning.}
    \label{fig:management_agent}
\end{figure*}

Unlike standard large language models that perform transient processing within restricted windows, agents implement persistent regulation of experiences via explicit management mechanisms. 
As illustrated in \autoref{fig:management_agent}, memory management serves as a cognitive operating system that establishes a closed-loop pipeline comprising extraction, updating, retrieval, and utilization. 
Within this cycle, extraction transforms interaction streams into structured records, while updating and retrieval ensure knowledge validity and precise localization, respectively. 
Utilization subsequently leverages these experiences for decision support. This framework enables agents to evolve from stateless responders into continuous learners capable of long-range reasoning. 
In the following, we sequentially examine these four critical stages.

\subsubsection{Memory Extraction}
\label{ssec:Memory Extraction in Agents}

Echoing the encoding and integration mechanisms in cognitive neuroscience, memory extraction in agents serves as the initial phase of memory management. 
Its core lies in distilling compact and meaningful content from the overwhelming and redundant information stream to construct usable memory representations. 
Based on the levels of information abstraction and processing methodologies, existing memory extraction methods can be broadly categorized into three paradigms, namely flat extraction, hierarchical extraction, and generative extraction.

\textbf{Flat Extraction.}
Flat extraction constitutes the foundational paradigm of memory construction, characterized by the direct recording of raw information into storage or the application of lightweight preprocessing such as summarization and segmentation~\cite{method_tremu, method_maq}. 
For instance, \citet{method_memgpt} drew inspiration from hierarchical storage concepts in operating systems, utilizing recall storage and archive storage to preserve complete dialogue histories and segmented document fragments, respectively, thereby constructing a virtual infinite context. 
Similarly, \citet{method_talm} deposited historical task execution data directly into long-term memory, facilitating cross-task experience reuse and reasoning enhancement through retrieval mechanisms. 
However, the direct storage of raw trajectories often incurs the risk of error accumulation. 
To address this, \citet{method_mga} proposed a decoupled extraction strategy that distilled continuous interaction streams into independent memory units across multiple dimensions, effectively blocking the propagation of local reasoning errors from the original long chains to subsequent decisions, while \citet{method_reasoningbank} focused on extracting high-level reasoning patterns from interaction trajectories to establish a scalable reasoning memory, thereby driving agent self-evolution.

\textbf{Hierarchical Extraction.}
Hierarchical extraction organizes fragmented information into ordered structures through multi-granular abstraction mechanisms, aiming to emulate the human cognitive ability to flexibly switch between macro-contexts and micro-details~\cite{method_beyond_retrieval, method_r2d2, method_smurfs, method_bridging_intuitive, method_nemori, method_agentfly}. 
In long-document processing, \citet{method_readagent, method_cam} employed recursive compression strategies, extracting high-level gist memory or constructing hierarchical semantic networks from raw text blocks, while indexing low-level details only when necessary to balance understanding depth and breadth. 
In interaction and dialogue scenarios, the extraction process typically adheres to the distinction between episodic and semantic memory. 
\citet{method_personaagent, method_memweaver, method_learning_from_supervision, method_mem_alpha} encoded interaction streams into specific episodic memory (e.g., timestamped events, graph-based behavioral trajectories) and abstracted semantic memory (e.g., user profiles, factual knowledge, or core summaries). 
This dual-coding mechanism was similarly applied by \citet{method_repository_memory} to extract code repository commit histories versus functional summaries. 
For complex task planning, extraction focuses on decoupling high-level planning from low-level execution. 
\citet{method_h2r, method_legomem} stored macroscopic instructions from planners separately from specific actions of executors, while \citet{method_muse} distinguished between high-level strategies (e.g., dilemma solutions) and low-level procedures (e.g., subtask standard operating procedures (SOPs)) to facilitate fine-grained knowledge transfer. 
Unlike methods relying on static predefined structures, \citet{method_a_mem} drew on the zettelkasten method, constructing a self-organizing dynamic hierarchical memory through active context linking and knowledge evolution.

\textbf{Generative Extraction.}
Diverging from the aforementioned paradigms that rely on external storage or hierarchical organization, generative extraction aims to dynamically reconstruct context during the reasoning process, thereby mitigating the computational overhead and attention dilution associated with excessive context length. 
\citet{method_mem1} enhanced context integration capabilities through training, compressing historical memory and reasoning into a shared representation space prior to each interaction, thereby facilitating continuous interaction without reliance on external memory modules. 
In contrast, \citet{method_resum} employed a specialized summarization model as a tool, which was actively triggered to condense think-action-observation trajectories whenever context usage approached a predefined threshold. 
Distinguishing itself from linear summarization approaches, \citet{method_foldgrpo} introduced branch and return operators, allowing agents to construct isolated sub-contexts for specific subtasks and inject only concise post-execution summaries back into the main context. 
Furthermore, \citet{method_agentfold} advanced this by training granular condensation and deep consolidation capabilities. 
This enables not only the extraction of summary chunks from individual interactions but also the recursive merging of multiple chunks into a single abstract representation when specific sequences become irrelevant (e.g., upon subtask completion or path abandonment), significantly compressing context size while preserving critical reasoning outcomes.

\subsubsection{Memory Updating}
\label{ssec:Memory Updating in Agents}
While memory extraction (\S\ref{ssec:Memory Extraction in Agents}) converts raw interaction streams into structured records, the resultant memory is not a static repository but a dynamic process of reconstruction. 
Echoing the mechanisms of reconsolidation and memory updating in cognitive neuroscience, memory updating in agents aims to balance the intake of newly extracted information with the elimination of the obsolete, ensuring system plasticity and efficiency. 
This process unfolds across two distinct layers: 
(1) Inside-Trial Updating, which focuses on the dynamic refreshing of the immediate context window (working memory) during a specific task execution, and 
(2) Cross-Trial Updating, which governs the lifecycle management of long-term archives (external memory) to persist knowledge across different episodes.

\textbf{Inside-Trial Updating.}
Inside-trial updating primarily operates on working memory to address the challenges of information decay and overload within limited context windows, ensuring high information density throughout continuous interactions~\cite{method_atomagents, method_taskgen, method_memllm, method_select_read_write, method_simpledoc, method_kg_agent}. 
Mimicking human selective attention, \citet{dy_memory_mem} proposed a filter-based update strategy that reshaped the context stream by real-time assessment, retaining only critical details while discarding irrelevant noise. 
In dynamic scenarios involving external tools, the focus of update shifts to just-in-time resource management.
\citet{method_memtool} introduced an “Add-Delete-Manage” strategy to dynamically remove irrelevant tools and retrieve new ones within a fixed window, balancing memory overhead with task success rates. 
Furthermore, \citet{method_sfr_deepresearch} treated memory updating as a cognitive operation actively invoked by the model, triggering summarization actions at specific junctures to compress lengthy historical trajectories into a refreshed, compact context. 
For more methodologies leveraging generative mechanisms for context updates, please refer to Generative Extraction (\S\ref{ssec:Memory Extraction in Agents}).

\textbf{Cross-Trial Updating.}
Cross-trial updating fundamentally represents the dynamic iteration and quality maintenance of the agent's external knowledge base, aiming to resolve the conflict between infinite knowledge expansion and limited storage capacity over time~\cite{method_user_behavior_simulation, method_ram, method_ever_evolving, method_rememberme_refineme}. 
To maintain the freshness and high value of knowledge, fundamental update strategies primarily focus on establishing mechanisms for selective retention and forgetting.
\citet{method_think_in_memory} eliminated redundant ideas through consistency maintenance within buckets.
\citet{cog_memory_sage,method_moom} introduced biologically inspired forgetting mechanisms, utilizing the Ebbinghaus forgetting curve~\cite{ebbinghaus} and competition-inhibition theory~\cite{theory_competition_inhibition}, respectively, to automatically phase out low-value or obsolete knowledge fragments, while \citet{dy_memory_light} ensured efficient knowledge retrieval through lightweight pruning strategies. 
Building on this to achieve deeper knowledge integration and structuring, \cite{method_tfgrpo} leverages semantic understanding to perform fine-grained refinement of the experience pool, removing the coarse and extracting the essential. 
\citet{method_flex} employed a dedicated updater agent to semantically merge new knowledge with existing entries and insert them into the correct hierarchical positions, ensuring the orderly expansion of the knowledge system. 
Furthermore, to enable knowledge update strategies to adapt to complex and changing task environments, \citet{method_memory_r1, method_mem_alpha} moved beyond static rules, utilizing reinforcement learning to train agents to autonomously explore optimal policies for knowledge retention and forgetting. 
Ultimately, \citet{method_open_ended} posited that knowledge updates should not be limited to passive adaptation but should be driven by autonomous goals set by the agent, thereby achieving true self-evolution through the active reconstruction of knowledge.

\subsubsection{Memory Retrieval}
\label{ssec:Memory Retrieval in Agents}
Once the knowledge base has been refined and maintained through Memory Updating (\S\ref{ssec:Memory Updating in Agents}), Memory Retrieval serves as the critical bridge between these retained experiences and dynamic decision-making.
Crucially, while retrieval in cognitive neuroscience is an active, transformative process that often involves reconsolidation to reshape the memory trace, retrieval in current agent architectures is primarily implemented as a selective activation mechanism. 
Driven by current contextual cues, this process selectively activates relevant information from the storage. 
By filtering irrelevant noise, it enables agents to leverage the vast knowledge repositories within limited context windows.
Based on the dimensions of retrieval strategies, existing methods are categorized into two paradigms: 
(1) Similarity-based Retrieval, which prioritizes semantic matching, 
and (2) Multi-factor Retrieval, which integrates multidimensional metrics such as recency and importance.

\textbf{Similarity-based Retrieval.}
Similarity-based retrieval represents the dominant paradigm in current agent memory retrieval, grounded in the core hypothesis that memory relevance is equivalent to semantic similarity.
Such approach typically utilizes encoders to map current queries into high-dimensional vectors, calculating cosine similarity within the vector space to recall the top-$k$ nearest fragments from storage. 
This retrieval-augmented generation (RAG)~\cite{ex_memory_rag} paradigm is widely adopted in question-answering and conversational agents, providing models with an efficient pathway for non-parametric knowledge access~\cite{method_licomemory}. 
However, while similarity-based retrieval excels in handling declarative knowledge, \citet{benchmark_procedural_memory} pointed out that direct embedding matching often fails to capture the structured operational logic inherent in procedural memory, thereby hindering agents from transferring complex behavioral strategies based solely on surface-level similarity.

\textbf{Multi-factor Retrieval.}
Multi-factor retrieval transcends the singular semantic dimension, moving beyond reliance solely on content similarity to determine memory prioritization based on factors such as recency, importance, structural efficiency, and expected rewards~\cite{method_understands_better, method_human_inspired, method_rsp, method_learn_to_memorize,method_assomem}. 
As a foundational work in this paradigm, \citet{method_generative_agents} established a composite scoring system synthesizing recency, importance, and relevance to simulate human memory decay and salience. 
Building on this, \citet{method_rcr_router} addressed resource-constrained multi-agent scenarios by introducing a token-budget-aware routing strategy, which calculates importance scores based on role relevance and task stage priority to distribute high-value information within limited windows. 
To resolve efficiency bottlenecks and fragmentation in global scanning, retrieval mechanisms have evolved towards structured approaches. 
For instance, \citet{method_h_mem} constructed a semantic hierarchical structure using top-down indexing to bypass expensive global computation, while \citet{method_multi_granularity} linked multi-granular representations via association graphs to dynamically retrieve memory at the most appropriate granularity. 
Furthermore, to align retrieval directly with decision quality, \citet{method_m_mdp} moved beyond superficial similarity matching by encoding historical outcomes into expected Q-value rewards, utilizing online soft Q-learning to retrieve exemplars that maximize expected utility.
Ultimately, \cite{method_memocue} elevated the retrieval paradigm from passive information querying to active cognitive guidance, unlocking latent memory through strategy-driven prompting, marking a shift in agent retrieval mechanisms towards functioning as human-like cognitive guides.

\subsubsection{Memory Application}
\label{ssec:Memory Application in Agents}
Following \textit{Memory Retrieval} (\S\ref{ssec:Memory Retrieval in Agents}), the focus of memory management shifts to its application in guiding behavior. 
Mirroring the brain's dual mechanisms of employing short-term memory for immediate reasoning and consolidating long-term experiences into cortical connections, memory application in agents falls into two primary paradigms, namely contextual augmentation and parameter internalization.

\textbf{Contextual Augmentation.}
Standard Retrieval-Augmented Generation paradigms typically concatenate retrieved text statically into prompts, treating memory as a passive reference~\cite{ex_memory_rag}.
However, effective utilization of agentic memory goes beyond simple concatenation~\cite{method_ufo, method_agentlite, method_hisim, method_magic, methoed_prep}.
It serves as a core mechanism for synthesizing fragmented information in long-document understanding, maintaining consistent personas for long-term personalization, and actively reusing past experiences for reasoning. 
To realize such dynamic cognitive capabilities, \citet{method_gam} introduced Just-in-Time compilation, building task-optimized contexts from lossless storage to resolve detail loss in long documents. 
Furthermore, \citet{method_mem1} compressed historical interactions into a shared representation space acting as a cognitive scratchpad, enabling deep synergy between memory and reasoning, while \citet{method_worldmm} extended this mechanism to multimodal video streams, adaptively integrating memories across temporal scales for long-horizon planning.

\textbf{Parameter Internalization.}
Drawing from lifelong learning~\cite{survey_lifelong_learning}, this paradigm consolidates explicit memory into implicit parameters. 
Initiating with retrieval efficiency, \citet{method_memverse} proposed distilling long-term memories into lightweight models for low-cost experience recall. 
This transformation of memory-to-model is subsequently leveraged to drive agent self-evolution.
For instance, \citet{method_digirl} utilized raw interaction history for reinforcement learning in open environments, while \citet{method_webrl} used perplexity to filter appropriate difficulty trajectories for more efficient adaptive training. 
Scaling from individual evolution to collective intelligence, \citet{method_seagent} employed a hierarchical distillation strategy to transfer trajectory experiences from multiple specialists to a generalist model. 
Finally, to move beyond mere imitation, \citet{method_early_experience} exploited the value of early exploration memories, using contrastive training between non-expert and expert paths to help models understand the causal superiority of expert decisions.

%% file: 7_Benchmark.tex
\section{Benchmarks for Agent Memory}
\label{sec:Benchmarks for Agent Memory}
In this section, we provide a comprehensive review of representative benchmarks for evaluating the memory capabilities of LLM-based agents. 
Based on the perspective of memory observation, we categorize existing benchmarks into two types:
(1) Semantic-oriented benchmarks (\S\ref{ssec:Benchmarks for Semantic-oriented}) focus on examining an agent's ability to maintain and update its internal state, encompassing not only memory retrieval tests but also higher-order cognitive dimensions such as dynamics and self-evolution.
(2) Episodic-oriented benchmarks (\S\ref{ssec:Benchmarks for Episodic-oriented}) are designed to evaluate an agent's competence in vertical domains such as web search, tool use, and environmental interaction, with an emphasis on assessing overall performance in complex application scenarios.
Additionally, we divide the evaluation attributes of both semantic-oriented and episodic-oriented benchmarks into three dimensions. 
Fidelity measures the agent’s ability to accurately retrieve and reproduce details from long-term contextual history without distortion. 
Dynamics evaluates memory update capabilities, testing whether agents can perform maintenance operations such as modifying outdated information or deleting errors to ensure consistent reasoning across evolving conversations. 
Generalization assesses whether stored memory can be translated into policies that adjust agent behavior and improve performance in new environments.

\input{tables/benchamrks}

\subsection{Benchmarks for Semantic-oriented}
\label{ssec:Benchmarks for Semantic-oriented}
Semantic-oriented benchmarks focus on how an agent constructs, maintains, and utilizes its internal memory state (e.g., fact and conception). 
These benchmarks exhibit a hierarchical evaluation structure, starting with the most basic memory capability, examining the agent's accurate retention and retrieval of information across multiple rounds of dialogue or long text narratives. 
Representative benchmarks include LoCoMo~\cite{bench_mem_locomo}, LOCCO~\cite{bench_mem_locco}, BABILong~\cite{bench_mem_babilong}, MPR~\cite{bench_mem_mpr}, RULER~\cite{bench_mem_ruler}, HotpotQA~\cite{bench_mem_hotpotqa}, PerLTQA~\cite{bench_mem_perltqa}, MemDaily~\cite{bench_mem_memdaily}.
These benchmarks target agents with external memory storage capabilities, evaluating their ability to accurately retrieve historical details and maintain semantic fidelity as memory span increases and distracting information accumulates, while avoiding information distortion caused by retrieval noise.

Some other benchmarks evaluate memory from a process perspective.
They focus on how information states are continuously updated and synchronized in real time as the environment evolves.
Representative benchmarks include MemBench~\cite{bench_dy_membench}, LongMemEval~\cite{bench_mem_longmemeval}, MemoryBank~\cite{bench_mem_memorybank}, DialSim~\cite{bench_mem_memdaily}, PrefEval~\cite{bench_mem_preeval}, SHARE~\cite{bench_dy_share}. 
These benchmarks require agents to proactively identify and discard outdated information while ensuring the persistence and consistency of valid information over long periods and across dialogues.

A further line of benchmarks measures how efficiently agents transform interaction history into cognitive abilities through self-evolutionary behaviors such as self-reflection, memory abstraction, and policy iteration.
Benchmarks such as LTMBenchmark~\cite{bench_evo_ltmb}, StoryBench~\cite{bench_evo_story}, MemoryAgentBench~\cite{bench_evo_mem}, Evo-Memory~\cite{bench_evo_evo}, MemBench~\cite{bench_dy_memorybench}, HaluMem~\cite{bench_dy_halumem}, LifelongAgentBench~\cite{bench_dy_lifelongagentbench}, StreamBench~\cite{bench_dy_stream}, StoryBench~\cite{bench_evo_storybench} provide experimental fields that tolerate trial and error and allow for continuous state changes for this type of research. 
These benchmarks require agents not only to maintain long-term memory consistency, but also to abstract general rules from past error patterns and transfer them to novel scenarios, thereby achieving meta-level cognitive improvement.

\subsection{Benchmarks for Episodic-oriented}
\label{ssec:Benchmarks for Episodic-oriented}
Unlike semantic-centric benchmarks that focus on the memory mechanism itself, episodic-oriented benchmarks aim to evaluate the actual performance gains of memory systems on agents in complex downstream application scenarios. 
Their core focus is on whether agents effectively leverage accumulated experience from memory to enhance performance on future tasks.
Through efficient information integration and state tracking, the memory empower agents to handle task flows in vertical domains such as web search, tool-use, and environmental interaction, thereby achieving higher levels of planning and execution in real-world scenarios with high time dependence and logical complexity.

\input{tables/benchmarks_other}

In the field of web search, task-oriented benchmarks have evolved from simple information retrieval to a comprehensive evaluation of an agent's autonomous navigation and interaction capabilities within dynamic web pages. 
Benchmarks such as WebChoreArena~\cite{bench_web_webchorearena}, WebArena~\cite{bench_web_arena}, WebShop~\cite{bench_web_webshop}, cover scenarios in simulated and real-world web environments, examining how agents utilize memory to maintain consistency in long-term tasks within highly dynamic and structurally complex web page interactions. 
The evaluation results of these benchmarks also demonstrate that efficient memory mechanisms are crucial for agents to ensure functional correctness and logical completeness in task flows.

The evaluation method for tool use has also been upgraded from atomic API calls to complex workflow reasoning. 
ToolBench~\cite{bench_tool_xbenchds}, GAIA~\cite{bench_tool_gaia}, xBench-DS~\cite{bench_tool_xbenchds}, benchmarks focus on examining how memory helps agents accurately retrieve tool schemas and maintain context state in multimodal and long-view tasks. 
Their evaluations have also demonstrated the key value of memory in tool use in overcoming execution illusions and establishing complex task logic through adaptive trial and error mechanisms.

Benchmarks for interaction with the environment focus on evaluating an agent’s perception-action mapping capabilities in dynamic and observable environments under different conditions, with a particular emphasis on the role of memory in supporting state tracking, causal inference, and personalized alignment. 
For example, BabyAI~\cite{bench_env_babyai} evaluated sample efficiency in course learning, particularly its ability to remember and combine sub-goals in long-sequence navigation.
ScienceWorld~\cite{bench_env_scienceworld} extended the scenario to scientific experiment simulations, requiring the agent to continuously track environmental variables using memory and verify causal inference capabilities through multi-step operations. 
Mind2Web~\cite{bench_env_mind2web} spanned heterogeneous web page tasks across multiple domains, enabling the agent to filter environmental noise and achieve cross-domain generalization when facing complex document object model (DOM) structures. 
PersonalWAB~\cite{bench_web_personalwab} further enriched the contextual dimension of the environment, incorporating user profiles and historical behavior into the interaction loop, and evaluated the accuracy of memory in achieving personalized intent alignment during dynamic interactions. 
AgentOccam~\cite{bench_env_AgentOccam} addressed the observation-action alignment problem in complex web environments, revealing that memory must possess the ability to prune and reconstruct massive amounts of environmental observations.

In summary, the practical significance of episodic-oriented benchmarks lies in establishing the agent's transformation from a conversationalist to an executor, and promoting the shift of the evaluation paradigm from dialogue-centric to problem-solving-centric. 
In summary, the practical significance of episodic-oriented benchmarks lies in establishing the agent's transformation from a conversationalist to an executor, and promoting the shift of the evaluation paradigm from model-centric to problem-solving-centric. 
By introducing noise from the real environment, it verifies the robustness of memory in maintaining state tracking under non-ideal conditions, thus bridging the gap between simulation and real-world applications.

%% file: tables/benchamrks.tex
\begin{table}[]
\centering
\resizebox{\textwidth}{!}{%
    \begin{tabular}{ccccccccc}
    \toprule 
    \multirow{2}{*}[-1ex]{\textbf{Benchmarks}} & \multicolumn{2}{c}{\textbf{Links}} & \multicolumn{3}{c}{\textbf{Attribute}} & \multicolumn{3}{c}{\textbf{Task}} \\ 
    \cmidrule(lr){2-3} \cmidrule(lr){4-6} \cmidrule(lr){7-9} 
     & \textbf{HomePage} & \textbf{GitHub} & \textbf{Fid.} & \textbf{Dyn.} & \textbf{Gen.} & \textbf{Task Type} & \textbf{Data Type} & \textbf{Data Quantity} \\ \midrule 
    LoCoMo \cite{bench_mem_locomo} & \href{https://snap-research.github.io/locomo/}{\faHome\ HomePage}  & \href{https://github.com/snap-research/LoCoMo}{\faGithub\ GitHub} & \scalebox{1.5}{\cmark} & \scalebox{1.5}{\xmark}& \scalebox{1.5}{\xmark} & Cross-Session Reasoning  & Text + Image & 50 \\
    MemoryAgentBench \cite{bench_evo_memoryagentbench} & \href{https://arxiv.org/abs/2507.05257}{\faHome\ HomePage}  & \href{https://github.com/HUST-AI-HYZ/MemoryAgentBench}{\faGithub\ GitHub} & \scalebox{1.5}{\cmark} & \scalebox{1.5}{\cmark}& \scalebox{1.5}{\cmark} & Memory Competency Evaluation  & Text & 2,071 \\
    MemBench \cite{bench_dy_membench} & \href{https://arxiv.org/abs/2506.21605}{\faHome\ HomePage}  & \href{https://github.com/import-myself/Membench}{\faGithub\ GitHub} & \scalebox{1.5}{\cmark} & \scalebox{1.5}{\cmark}& \scalebox{1.5}{\xmark} & Multi-level Memory Evaluation  & Text  & 65,000  \\
    LongMemEval \cite{bench_mem_longmemeval} & \href{https://xiaowu0162.github.io/long-mem-eval/}{\faHome\ HomePage}  & \href{https://github.com/xiaowu0162/LongMemEval}{\faGithub\ GitHub} & \scalebox{1.5}{\cmark} & \scalebox{1.5}{\cmark}& \scalebox{1.5}{\xmark} & Long-term Memory Abilities  & Text & 500 \\
    MemoryBank \cite{bench_mem_memorybank} & \href{https://arxiv.org/abs/2305.10250}{\faHome\ HomePage}  & \href{https://github.com/zhongwanjun/MemoryBank-SiliconFriend}{\faGithub\ GitHub} & \scalebox{1.5}{\cmark} & \scalebox{1.5}{\cmark}& \scalebox{1.5}{\xmark} & Long-term AI Companion Evaluation  & Text & 194  \\
    MemoryBench \cite{bench_dy_memorybench} & \href{https://arxiv.org/abs/2510.17281}{\faHome\ HomePage}  & \href{https://github.com/LittleDinoC/MemoryBench}{\faGithub\ GitHub} & \scalebox{1.5}{\cmark} & \scalebox{1.5}{\cmark}& \scalebox{1.5}{\cmark} & Continual Learning  & Text & 20,000  \\
    HaluMem \cite{bench_dy_halumem} & \href{https://arxiv.org/abs/2511.03506}{\faHome\ HomePage}  & \href{https://github.com/MemTensor/HaluMem}{\faGithub\ GitHub} & \scalebox{1.5}{\cmark} & \scalebox{1.5}{\cmark}& \scalebox{1.5}{\cmark} & Hallucination Evaluation in Memory  & Text  & 83,000 \\
    Evo-Memory \cite{bench_evo_evo} & \href{https://arxiv.org/abs/2511.20857}{\faHome\ HomePage}  & \href{https://github.com/devopsdymyr/Evo-Memory}{\faGithub\ GitHub} & \scalebox{1.5}{\cmark} & \scalebox{1.5}{\cmark}& \scalebox{1.5}{\cmark} & Self-Evolving memory & Text & -  \\
    LTMBenchmark \cite{bench_evo_ltmb} & \href{https://arxiv.org/abs/2409.20222}{\faHome\ HomePage}  & \href{https://github.com/GoodAI/goodai-ltm-benchmark}{\faGithub\ GitHub} & \scalebox{1.5}{\cmark} & \scalebox{1.5}{\cmark}& \scalebox{1.5}{\cmark} & Dynamic conversation  & Text  & 33     \\
    LOCCO \cite{bench_mem_locco} & \href{https://aclanthology.org/2025.findings-acl.1014/}{\faHome\ HomePage}  & \href{https://github.com/JamesLLMs/LoCoGen}{\faGithub\ GitHub} & \scalebox{1.5}{\cmark} & \scalebox{1.5}{\xmark}& \scalebox{1.5}{\xmark} & Long-term Memory Abilities  & Text & 3,080 \\
    PersonaMem \cite{bench_mem_personamem} & \href{https://arxiv.org/abs/2504.14225}{\faHome\ HomePage}  & \href{https://github.com/bowen-upenn/PersonaMem}{\faGithub\ GitHub} & \scalebox{1.5}{\cmark} & \scalebox{1.5}{\cmark}& \scalebox{1.5}{\cmark} & Dynamic User Profiling and Personalization  & Text  & 180 \\
    LongBench \cite{bench_mem_longbench} & \href{https://arxiv.org/abs/2308.14508}{\faHome\ HomePage}  & \href{https://github.com/THUDM/LongBench}{\faGithub\ GitHub} & \scalebox{1.5}{\cmark} & \scalebox{1.5}{\xmark}& \scalebox{1.5}{\cmark} & long-text application  & Text  & 4,750 \\
    LongBench v2 \cite{bench_mem_longbenchv2} & \href{https://longbench2.github.io/}{\faHome\ HomePage}  & \href{https://github.com/THUDM/LongBench}{\faGithub\ GitHub} & \scalebox{1.5}{\cmark} & \scalebox{1.5}{\xmark}& \scalebox{1.5}{\cmark} & Deep Reasoning and Realistic Multitasks   & Text & 503 \\
    BABILong \cite{bench_mem_babilong} & \href{https://arxiv.org/abs/2406.10149}{\faHome\ HomePage}  & \href{https://github.com/booydar/babilong}{\faGithub\ GitHub} & \scalebox{1.5}{\cmark} & \scalebox{1.5}{\xmark}& \scalebox{1.5}{\xmark} & Reasoning-in-a-Haystack  & Text & 13,000 \\
    RULER \cite{bench_mem_ruler} & \href{https://arxiv.org/abs/2404.06654}{\faHome\ HomePage}  & \href{https://github.com/NVIDIA/RULER}{\faGithub\ GitHub} & \scalebox{1.5}{\cmark} & \scalebox{1.5}{\xmark}& \scalebox{1.5}{\xmark} & Synthetic Long-Context Evaluation  & Text  & 39,000  \\
    MM-Needle \cite{bench_mem_mmneedle} & \href{https://mmneedle.github.io/}{\faHome\ HomePage}  & \href{https://github.com/Wang-ML-Lab/multimodal-needle-in-a-haystack}{\faGithub\ GitHub} & \scalebox{1.5}{\cmark} & \scalebox{1.5}{\xmark}& \scalebox{1.5}{\xmark} & Multimodal Needle-in-a-Haystack  & Text + Image  & 40,000 \\
    DialSim \cite{bench_dy_dialsim} & \href{https://arxiv.org/abs/2406.13144}{\faHome\ HomePage}  & \href{}{-} & \scalebox{1.5}{\cmark} & \scalebox{1.5}{\cmark}& \scalebox{1.5}{\xmark} & Multi-party Dialogue Understanding  & Text & 1,300 \\
    LifelongAgentBench \cite{bench_dy_lifelongagentbench} & \href{https://arxiv.org/abs/2505.11942}{\faHome\ HomePage}  & \href{https://github.com/caixd-220529/LifelongAgentBench}{\faGithub\ GitHub} & \scalebox{1.5}{\cmark} & \scalebox{1.5}{\cmark}& \scalebox{1.5}{\cmark} & Lifelong learning  & Text & 1,400  \\
    PrefEval \cite{bench_mem_preeval} & \href{https://prefeval.github.io/}{\faHome\ HomePage}  & \href{https://github.com/amazon-science/PrefEval}{\faGithub\ GitHub} & \scalebox{1.5}{\cmark} & \scalebox{1.5}{\cmark}& \scalebox{1.5}{\xmark} & Personalized Preference Following  & Text  & 3,000  \\
    MPR \cite{bench_mem_mpr} & \href{https://arxiv.org/abs/2508.13250}{\faHome\ HomePage}  & \href{https://github.com/nuster1128/MPR}{\faGithub\ GitHub} & \scalebox{1.5}{\cmark} & \scalebox{1.5}{\xmark}& \scalebox{1.5}{\xmark} & Multi-hop Personalized Reasoning  & Text  & 108,000 \\
    StreamBench \cite{bench_dy_stream} & \href{https://stream-bench.github.io/}{\faHome\ HomePage}  & \href{https://github.com/stream-bench/stream-bench}{\faGithub\ GitHub} & \scalebox{1.5}{\cmark} & \scalebox{1.5}{\cmark}& \scalebox{1.5}{\cmark} & Continuous Online Learning  & Text & 9,702 \\
    Madial-Bench \cite{bench_mem_madialbench} & \href{https://aclanthology.org/2025.naacl-long.499/}{\faHome\ HomePage}  & \href{https://github.com/hejunqing/MADial-Bench}{\faGithub\ GitHub} & \scalebox{1.5}{\cmark} & \scalebox{1.5}{\xmark}& \scalebox{1.5}{\xmark} & Memory-Augmented  Dialogue  & Text  & 160 \\
    HotpotQA \cite{bench_mem_hotpotqa} & \href{https://hotpotqa.github.io/}{\faHome\ HomePage}  & \href{https://github.com/hotpotqa/hotpot}{\faGithub\ GitHub} & \scalebox{1.5}{\cmark} & \scalebox{1.5}{\xmark}& \scalebox{1.5}{\xmark} & Multi-hop Question Answering  & Text & 112,779 \\     
    PerLTQA \cite{bench_mem_perltqa} & \href{https://arxiv.org/abs/2402.16288}{\faHome\ HomePage}  & \href{https://github.com/Elvin-Yiming-Du/PerLTQA}{\faGithub\ GitHub} & \scalebox{1.5}{\cmark} & \scalebox{1.5}{\xmark}& \scalebox{1.5}{\xmark} & The use of personalized memory  & Text & 8,593  \\
    StoryBench \cite{bench_evo_storybench} & \href{https://arxiv.org/abs/2506.13356}{\faHome\ HomePage}  & \href{}{ - } & \scalebox{1.5}{\cmark} & \scalebox{1.5}{\cmark}& \scalebox{1.5}{\cmark} & Interactive fiction games  & Text  & -  \\ 
    SHARE \cite{bench_dy_share} & \href{https://aclanthology.org/2025.acl-long.704/}{\faHome\ HomePage}  & \href{https://github.com/e1kim/share/}{\faGithub\ GitHub} & \scalebox{1.5}{\cmark} & \scalebox{1.5}{\cmark}& \scalebox{1.5}{\xmark} & Skill learning and Transfer & Text  & 1,201  \\
    MovieChat-1K \cite{bench_env_moviechat1k} & \href{https://wenhaochai.com/MovieChat/}{\faHome\ HomePage}  & \href{https://github.com/rese1f/MovieChat}{\faGithub\ GitHub} & \scalebox{1.5}{\cmark} & \scalebox{1.5}{\cmark}& \scalebox{1.5}{\xmark} & Long Video Comprehension and QA & Text + Movie  & 15,000 \\
    Mobile-Agent-E \cite{bench_env_mobileagente} & \href{https://x-plug.github.io/MobileAgent/}{\faHome\ HomePage}  & \href{https://github.com/X-PLUG/MobileAgent}{\faGithub\ GitHub} & \scalebox{1.5}{\cmark} & \scalebox{1.5}{\cmark}& \scalebox{1.5}{\cmark} &  Complex mobile tasks & Text + Image & 25 \\   
    M3-Agent \cite{bench_env_m3bench} & \href{https://m3-agent.github.io/}{\faHome\ HomePage}  & \href{https://github.com/bytedance-seed/m3-agent}{\faGithub\ GitHub} & \scalebox{1.5}{\cmark} & \scalebox{1.5}{\cmark}& \scalebox{1.5}{\xmark} & Long-form video QA and memory reasoning  & Text + Audio +Video  & 5,510  \\
    Text2Mem Bench \cite{bench_other_text2mem} & \href{https://arxiv.org/abs/2509.11145}{\faHome\ HomePage}  & \href{https://github.com/MemTensor/text2mem}{\faGithub\ GitHub} & \scalebox{1.5}{\cmark} & \scalebox{1.5}{\cmark}& \scalebox{1.5}{\xmark} & Memory operation instruction execution  & Text & - \\
    SoMoSiMu-Bench \cite{bench_env_somosimubench} & \href{https://arxiv.org/abs/2402.16333}{\faHome\ HomePage}  & \href{https://github.com/xymou/social_simulation}{\faGithub\ GitHub} & \scalebox{1.5}{\cmark} & \scalebox{1.5}{\cmark}& \scalebox{1.5}{\xmark} & Social Movement Simulation  & Text + Structured data & 6,700  \\
     MemDaily \cite{bench_mem_memdaily} & \href{https://arxiv.org/abs/2409.20163}{\faHome\ HomePage}  & \href{https://github.com/nuster1128/MemSim}{\faGithub\ GitHub} & \scalebox{1.5}{\cmark} & \scalebox{1.5}{\xmark}& \scalebox{1.5}{\xmark} & Personalized Reasoning  & Text  & 26,003  \\
    \bottomrule 
    \end{tabular}%
}
\caption{Representative benchmarks for evaluating semantic-oriented agents. The attributes are defined as follows: \textbf{Fid. (Fidelity)} measures the agent's ability to accurately retrieve and reproduce information from long contexts. \textbf{Dyn. (Dynamics)} assesses whether the memory system supports dynamic updates (e.g., modification, deletion) and cross-session reasoning. \textbf{Gen. (Generalization)} evaluates whether the stored memory can evolve into strategies that adapt the agent's behavior in novel environments.}
\end{table}

%% file: tables/benchmarks_other.tex
\begin{table}[]
\centering
\resizebox{\textwidth}{!}{%
    \begin{tabular}{ccccccccc}
    \toprule 
    \multirow{2}{*}[-1ex]{\textbf{Benchmarks}} & \multicolumn{2}{c}{\textbf{Links}} & \multicolumn{3}{c}{\textbf{Attribute}} & \multicolumn{3}{c}{\textbf{Task}} \\ 
    \cmidrule(lr){2-3} \cmidrule(lr){4-6} \cmidrule(lr){7-9} 
     & \textbf{HomePage} & \textbf{GitHub} & \textbf{Fid.} & \textbf{Dyn.} & \textbf{Gen.} & \textbf{Task Type} & \textbf{Data Type} & \textbf{Data Quantity} \\ \midrule 
    WebChoreArena \cite{bench_web_webchorearena} & \href{https://arxiv.org/abs/2506.01952}{\faHome\ HomePage}  & \href{https://github.com/WebChoreArena/WebChoreArena}{\faGithub\ GitHub} & \scalebox{1.5}{\cmark} & \scalebox{1.5}{\xmark}& \scalebox{1.5}{\xmark} & Massive memory + Calculation & Text + Image  & 532 \\
    ScienceWorld \cite{bench_env_scienceworld} & \href{https://sciworld.apps.allenai.org/}{\faHome\ HomePage}  & \href{https://github.com/allenai/ScienceWorld}{\faGithub\ GitHub} & \scalebox{1.5}{\cmark} & \scalebox{1.5}{\cmark}& \scalebox{1.5}{\cmark} & Scientific Experiments and Reasoning & Text & 7,200  \\
    WebArena \cite{bench_web_arena} & \href{https://webarena.dev/}{\faHome\ HomePage}  & \href{https://github.com/web-arena-x/webarena}{\faGithub\ GitHub} & \scalebox{1.5}{\cmark} & \scalebox{1.5}{\cmark}& \scalebox{1.5}{\xmark} & Simulating human web operations & Text + Image & 812  \\
    WebShop \cite{bench_web_webshop} & \href{https://webshop-pnlp.github.io/}{\faHome\ HomePage}  & \href{https://github.com/princeton-nlp/WebShop}{\faGithub\ GitHub} & \scalebox{1.5}{\cmark} & \scalebox{1.5}{\cmark}& \scalebox{1.5}{\xmark} & E-commerce shopping navigation and decision  & Text + Image & 12,087 \\
    ToolBench \cite{bench_tool_toolbench} & \href{https://arxiv.org/abs/2307.16789}{\faHome\ HomePage}  & \href{https://github.com/OpenBMB/ToolBench}{\faGithub\ GitHub} & \scalebox{1.5}{\cmark} & \scalebox{1.5}{\cmark}& \scalebox{1.5}{\xmark} & Tool use and Api call & Text & 142,950  \\
    GAIA \cite{bench_tool_gaia} & \href{https://arxiv.org/abs/2311.12983}{\faHome\ HomePage}  & \href{https://huggingface.co/gaia-benchmark}{\faGithub\ GitHub} & \scalebox{1.5}{\cmark} & \scalebox{1.5}{\cmark}& \scalebox{1.5}{\xmark} & General AI Assistants & Text + Image + Spreadsheet  & 466 \\
    BabyAI \cite{bench_env_babyai} & \href{https://arxiv.org/abs/1810.08272}{\faHome\ HomePage}  & \href{https://github.com/mila-iqia/babyai}{\faGithub\ GitHub} & \scalebox{1.5}{\cmark} & \scalebox{1.5}{\cmark}& \scalebox{1.5}{\xmark} & Language Learning  & Text + symbolic observation & 100,000+ \\
    xBench-DS \cite{bench_tool_xbenchds} & \href{https://xbench.org/}{\faHome\ HomePage}  & \href{https://github.com/xbench-ai/xbench-evals}{\faGithub\ GitHub} & \scalebox{1.5}{\cmark} & \scalebox{1.5}{\xmark}& \scalebox{1.5}{\xmark} & Business Recruiting and Marketing  & Text  & 100 \\
    AgentOccam \cite{bench_env_AgentOccam} & \href{https://arxiv.org/abs/2410.13825}{\faHome\ HomePage}  & \href{https://github.com/amazon-science/AgentOccam?tab=readme-ov-file}{\faGithub\ GitHub} & \scalebox{1.5}{\cmark} & \scalebox{1.5}{\cmark}& \scalebox{1.5}{\xmark} & General Web Interaction  & Text + Image & 812 \\
    StuLife \cite{bench_env_stulife} & \href{https://ecnu-icalk.github.io/ELL-StuLife/}{\faHome\ HomePage}  & \href{https://github.com/ECNU-ICALK/ELL-StuLife}{\faGithub\ GitHub} & \scalebox{1.5}{\cmark} & \scalebox{1.5}{\cmark}& \scalebox{1.5}{\cmark} & Shared Memory-Aware & Text & 1,284 \\
    Mind2Web \cite{bench_env_mind2web} & \href{https://osu-nlp-group.github.io/Mind2Web/}{\faHome\ HomePage}  & \href{https://github.com/OSU-NLP-Group/Mind2Web}{\faGithub\ GitHub} & \scalebox{1.5}{\cmark} & \scalebox{1.5}{\cmark}& \scalebox{1.5}{\xmark} & General Web Agent Task & Text + Image + File & 2,350  \\
    PersonalWAB \cite{bench_web_personalwab} & \href{https://hongrucai.github.io/PersonalWAB/}{\faHome\ HomePage}  & \href{https://github.com/HongruCai/PersonalWAB}{\faGithub\ GitHub} & \scalebox{1.5}{\cmark} & \scalebox{1.5}{\cmark}& \scalebox{1.5}{\xmark} & Personalized Web Agent Tasks  & Text + Web API  & 9,000  \\ \bottomrule 
    \end{tabular}%
}
\caption{Representative benchmarks for evaluating episodic-oriented agent memory. The attributes are defined as follows: \textbf{Fid. (Fidelity)} measures the agent's ability to accurately retrieve and reproduce information from long contexts. \textbf{Dyn. (Dynamics)} assesses whether the memory system supports dynamic updates (e.g., modification, deletion) and cross-session reasoning. \textbf{Gen. (Generalization)} evaluates whether the stored memory can evolve into strategies that adapt the agent's behavior in novel environments.}
\end{table}

%% file: 8_Security.tex
\section{Security of Agent Memory}
\label{sec:Security}
As LLM-driven agents are increasingly deployed for long-term task planning and tool utilization, memory has emerged as a critical component for maintaining state continuity. 
However, this centrality also renders memory a significant attack surface, with its associated security vulnerabilities becoming ever more pronounced~\cite{secu_memory_the,secu_memory_con}.
Specifically, we systematically examined the attack paradigm targeting agent memory (\S\ref{ssec:Attack of Agent Memory}) and summarized its defense paradigm (\S\ref{ssec:Defense of Agent Memory}).

\subsection{Attack of Agent Memory}
\label{ssec:Attack of Agent Memory}
In agent systems, memory is not only a repository of historical interaction information but also a crucial source of context for decision support. 
While this paradigm can improve the efficiency and capabilities of agents, it also exposes many security vulnerabilities. 
Attackers can exploit memory as a channel for privacy leaks or as a vehicle for implanting malicious logic. 
Specifically, we categorize attacks targeting agent memory into two types: 
(1) Extraction-based attack (\S\ref{sssec:Extraction-based Attack}) aims to steal user data from the memory,
and (2) Poisoning-based attack (\S\ref{sssec:Poisoning-based Attack}) compromises agent decision-making by injecting toxic information into the memory bank.

\subsubsection{Extraction-based Attack} 
\label{sssec:Extraction-based Attack}
In the category of data-targeted attacks, extraction-based attacks focus on stealing sensitive information from external data sources~\cite{method_unveiling}. 
As retrieval mechanisms evolve from basic models to complex agents, related research explains the gradual expansion of this attack surface. 
First, research on retrieval models such as KNN-LM~\cite{secu_extra_knn} showed that while introducing external private databases improves utility, it significantly increases the risk of privacy leakage, making it a weakness for attackers to reconstruct the original text. 
Based on this, for RAG, \citet{secu_extra_rag} proposed a composite structured prompting attack method consisting of information and command, aiming to quantify the privacy leakage risk of external retrieval databases in RAG, and at the same time verify the mitigation effect of the RAG mechanism on the privacy leakage of training data of large models. 
Further, for agents with complex workflows, \citet{secu_extra_unv} proposed a black-box attack framework containing specific attack prompts with locators and aligners, and combined with an LLM-based automated prompt generation strategy, successfully induced the agent to output sensitive user interaction history stored in its long-term memory.

\subsubsection{Poisoning-based Attack} 
\label{sssec:Poisoning-based Attack}
Poisoning-based attacks refer to attacks that do not require modification of model parameters but instead inject adversarially optimized malicious data into an external memory~\cite{method_prompt_infection}. 
These attacks exploit retrieval mechanism preferences and the model's excessive reliance on context to covertly implant backdoors or hijack the agent's decision-making logic. 
Therefore, we categorize poisoning-based attacks into highly concealed backdoor attacks.
Backdoor attacks constitute advanced persistent threats in agent memory security, with their effectiveness hinging on stealthy infiltration and precise triggering mechanisms.
Attackers inject malicious content into the retrieval database through carefully optimized triggers, enabling the agent to function normally under routine operations while executing adversarial behaviors only when specific trigger conditions are met.
\citet{secu_poi_poi,secu_poi_tro} both showed that attackers can precisely control the agent’s decision-making by manipulating the retrieval weights of memory in the vector space, and make retrieval the driver of backdoors.
In addition, attackers exploit the vulnerability of instruction following rather than relying on complex model training to disguise malicious instructions as ordinary memories and store them in the memory bank. 
\citet{secu_poi_not} demonstrated the vulnerability of the Agent when reading untrusted external data. 
\citet{secu_poi_in} confirmed that attackers do not need backend privileges, but can induce the agent to generate and store malicious memories and malicious records by simply querying and observing the agent's interactions with the agent, and can hijack subsequent query processing by using Bridging steps.
Furthermore, some poisoning attacks target the agent's cognition. The goal of such attacks is to inject large amounts of noise, conflicting information, or social biases into the memory bank, causing the agent's judgment to deteriorate or its values to become distorted \cite{secu_cog_cyb}.
\citet{secu_cog_dru} vividly described the process of making a recommendation agent ineffective, like a person who is drunk, by injecting confusing data. 
\citet{secu_cog_ra} established hidden associations by poisoning, causing the agent to continuously output discriminatory content.

\subsection{Defense of Agent Memory}
\label{ssec:Defense of Agent Memory}
Faced with diverse attack threats to agent memory, a series of defense mechanisms have been proposed in recent years, aiming to enhance the security and robustness of Agent Memory from different dimensions. 
Existing defense mechanisms have established a multi-layered defense system around three key stages, namely retrieval, response, and storage.
Specifically, the defense mechanism is divided into three key lines of defense: 
(1) Retrieval-based defense (\S\ref{sssec:Retrieval-based Defense}) responsible for purifying the system at its source,
(2) Response-based defense (\S\ref{sssec:Response-based Defense}) responsible for immediate blocking,
and (3) Privacy-based defense (\S\ref{sssec:Privacy-based Defense}) for protecting the complex underlying data. 
These three mechanisms complement each other, forming a closed-loop security defense system.

\subsubsection{Retrieval-based Defense}
\label{sssec:Retrieval-based Defense}
Retrieval-based defenses constitute the first line of defense for agent memory security. 
Their core logic lies in purifying the memory at its source, proactively blocking the propagation path of poisoning attacks before the retrieved external knowledge is integrated into explicit memory~\cite{secu_retri_sur}. 
Attackers manipulate retrieval results by implanting data poisoning into the memory. 
Defense strategies at this stage focus on textual features and contextual consistency, implementing anomaly detection and robust filtering. 
\citet{secu_retri_mem} proposed a consensus-based validation mechanism that constructs parallel reasoning paths by retrieving multiple related memories and uses the structural consensus formed by benign memories to identify and eliminate poisoning records that cause deviations in reasoning logic. 
\citet{secu_retri_gua} proposed a dual-agent defense framework to detect and repair poisoned chain-of-thought steps in code generation, ensuring the purity of retrieved context.

\subsubsection{Response-based Defense}
\label{sssec:Response-based Defense}
Response-based defense acts as the agent's cognitive immune system, ensuring that the agent will block the implementation of malicious logic even if the agent ingests memory fragments containing malicious instructions~\cite{secu_res_tiny}. 
For instance, \citet{secu_res_auto} proposed a multi-agent defense framework, which coordinates the input agent to make security presets, the Defense Agency to conduct collaborative review, and finally the output agent to decide how to output the final response to the user's request. 
\citet{secu_res_lats} integrated Monte Carlo Tree Search and self-reflection to rehearse and score multiple potential action trajectories in the response generation stage, avoiding high-risk paths induced by false memories or malicious intent.

\subsubsection{Privacy-based Defense}
\label{sssec:Privacy-based Defense}
Privacy-based defenses form the underlying safeguards for the data lifecycle, focusing on addressing the leakage and forgetting of sensitive information during memory retrieval. 
At its core, it ensures that sensitive data resides only in isolated private memory, using only anonymized or task-necessary memory for reasoning and collaboration, thereby preventing the leakage of private data during interactions~\cite{secu_pri_su,method_collaborative_memory}. 
For example, \citet{secu_pri_gama} proposed a general anonymizing multi-Agent system, which divides the workspace into private and public spaces. 
It uses domain rule-based knowledge enhancement and proof by contradiction-based logic enhancement to compensate for the semantic privacy caused by anonymization, thereby achieving efficient task processing and reasoning while ensuring privacy. 
\citet{secu_res_auto} proposed a context-integrity-based framework that uses a lightweight model to analyze user intent, automatically distinguishes and removes unnecessary sensitive information, and reconstructs prompts, thereby removing context-independent privacy data while preserving task intent.

%% file: 9_Future.tex
\section{Future Discussion}
\label{sec:Future}
As the field of memory research continues to advance, our focus shifts toward the next frontier of exploration. 
We explore promising research directions, particularly memory systems for multimodal agents (\S\ref{ssec:Multimodal Memory}) and agent skills that enable shared memory across agents (\S\ref{ssec:Agent Skills}).

\subsection{Multimodal Memory}
\label{ssec:Multimodal Memory}
Real-world environments inherently provide information beyond textual signals, involving diverse modalities such as vision, audio, and depth~\cite{survey_ustc,survey_hit,method_mm_vismem,method_mm_videomem,method_mm_agentic,method_mm_context_as_memory,method_mm_streaming_video}. 
Accordingly, agents deployed in complex settings typically perceive and interact with the environment through multimodal observations, with text being only one component. 
This has motivated growing interest in multimodal memory, extending beyond traditional text-only agent memory paradigms~\cite{survey_cuhk,method_augustus}.

Compared with pure text information, multimodal information is less well structured and contains more noise. 
Directly storing raw multimodal information often leads to memory waste and performance degradation~\cite{survey_cuhk}. 
DoraemonGPT~\cite{method_doraemongpt} and LifelongMemory~\cite{method_lifelongmemory} used expert models to transform raw multimodal information into structured and concise symbolic memories (e.g., timestamps, frame or clip-level captions~\cite{method_lavila}, object categories~\cite{method_ground_dino}). 
And MovieChat~\cite{method_moviechat} and MA-LLM~\cite{method_ma_llm} instead employed feature-level consolidation and compression of raw multimodal representations, using techniques such as token merging~\cite{method_token_merging}and Q-Former~\cite{method_blip2}. 
In addition, some works~\cite{method_jarvis_1,method_videoagent,method_m3agent,method_embodied_videoagent} simultaneously leveraged symbolic memories alongside their aligned multimodal content as evidence, resulting in a hybrid memory representation. 

Existing agent systems demonstrate that incorporating multimodal memory consistently improves performance across a wide range of environments. 
JARVIS-1~\cite{method_jarvis_1} enabled agents to perceive multimodal inputs and generate complex plans in interactive game environments like Minecraft.
\citet{method_flash_vstream,method_ma_llm} addressed the challenges of long contexts caused by dense video frames through efficient compression and storage, thereby enhancing the understanding of long videos. 
InternLM-XComposer2.5-OmniLive~\cite{method_internlm_xcomposer} and Video-SALMONN-S~\cite{method_video_salmonn_s} further integrated audio memory and visual memory to support online comprehension of audio–visual streams. 
M3-Agent~\cite{method_m3agent} and EgoMem~\cite{method_egomem} constructed and continuously updated entity-centric episodic and semantic memories from multimodal dialogues, enabling strong lifelong personalization capabilities. 
In application scenarios, \citet{method_appagentx,method_gui_agents} preserved the historical trajectories of GUI interaction states, allowing agents to identify repetitive operations and improve efficiency.

Despite some progress in recent years, existing methods still have limitations in several important aspects. 
Developing memory representations and operations that can seamlessly adapt to different modalities while ensuring semantic consistency and temporal alignment remains an unsolved problem. 
Furthermore, semantic degradation due to compression or abstraction, as well as unresolved issues such as long-term temporal dependency modeling and computational efficiency, still hinder the scalability of systems. 
Overcoming these challenges is crucial for building general-purpose intelligent agent systems capable of operating in complex multimodal environments.

\subsection{Agent Skills}
\label{ssec:Agent Skills}

Contemporary AI agents generally exhibit well-designed workflows, resulting in robust general-purpose capabilities. 
When augmented with memory systems, these capabilities are further enhanced, underscoring the pivotal role of memory in amplifying agent performance. 
However, as agents become increasingly sophisticated, there is a growing imperative to equip them with domain-specific expertise to accommodate a broader spectrum of application scenarios. 
For instance, \citet{method_nirvana} introduced a memory trigger mechanism that enables self-supervised adaptation during the test time, incorporating fast parameters to address tasks across diverse specialized domains. 
Nevertheless, this approach is relatively complex, and the learned parameters remain confined to the individual agent, precluding the transfer of specialized knowledge between agents. 
This phenomenon is akin to reinventing the wheel in isolation, squandering a wealth of invaluable experiential knowledge. 
Consequently, there is an urgent need for memory strategies that offer greater composability, scalability, and portability.

Agent skills~\cite{method_agentskills} is a modular capability extension paradigm proposed by Anthropic, with its core principle being the encapsulation of instruction sets, executable scripts, and associated resources into structured directory units. 
AI agents can dynamically discover, load, and execute these skill modules at runtime to accomplish domain-specific tasks. 
This mechanism abstracts domain expertise into composable and reusable modular resources, significantly expanding the capability boundaries of agents and enabling general-purpose agents to transform on demand into specialized agents tailored for vertical application scenarios. 
These modules function analogously to equipment in video games, where players equip corresponding gear when they seek particular attributes, and crucially, such equipment can be freely shared and transferred among players. 
Currently, several research~\cite{method_text2mem,method_agentkb,method_memengine} efforts have begun to explore this direction. 
For instance, \citet{method_text2mem} proposed a unified memory operation language aimed at providing standardized memory management interfaces for memory operating systems, thereby enabling portability and interoperability of memory representations. 
\citet{method_agentkb} constructed a unified knowledge management platform that allows intelligent agents across heterogeneous model architectures to access and manipulate shared memory resources. 
Although these works demonstrate the potential of this research direction, the field remains in its nascent stage, with numerous critical challenges awaiting further investigation:
(1) Unified Storage and Representation of Multimodal Information: Current memory systems are predominantly designed for textual modalities. 
How to construct a unified storage framework that supports multimodal information encompassing text, images, audio, and video, while simultaneously designing cross-modal retrieval and reasoning mechanisms, remains an open research question,
and (2) Cross-agent skill transfer and adaptation mechanisms: Different agent architectures, such as those built upon distinct foundation models, exhibit variations in capability characteristics and interface specifications. 
Designing a universal skill description language along with an adaptation layer that enables skill modules to be seamlessly transferred and reused across heterogeneous agents constitutes a critical challenge for realizing a genuine skill-sharing ecosystem.

%% file: 10_Conclusion.tex
\section{Conclusion}
\label{sec:conclusion}

In this comprehensive survey, we present an in-depth unified investigation of memory in both cognitive neuroscience and autonomous agents, covering definitions, functions, taxonomies, management, security, and future research directions. Although human-like memory mechanisms have proven to be highly beneficial for agents, existing work has yet to deeply draw upon the essence of memory mechanisms from brain science due to the gap between disciplines. We envision this survey as a step toward encouraging researchers to engage in cross-disciplinary collaboration and develop more robust agent memory systems.